\documentclass[11pt,a4paper]{article}
\usepackage{acl}
\usepackage{times}
\usepackage{nert}
\usepackage{latexsym}
\usepackage[T1]{fontenc}
\usepackage{bbm}
\usepackage{graphicx}
\usepackage{url}
\usepackage{multirow}
\usepackage{amsmath}
\usepackage{todonotes}
\usepackage{appendix}
\usepackage{mdframed}
\usepackage{makecell}
\usepackage{hhline}
\usepackage{subcaption}

\DeclareMathOperator*{\argmaxA}{arg\,max}

\usepackage{amsfonts}
\usepackage{xspace}
\usepackage{placeins}
\usepackage{soul}
\usepackage{caption}
\usepackage{newfloat}
\usepackage{tikz}
\definecolor{LimeGreen}{RGB}{179,253,148}
\definecolor{DarkLimeGreen}{RGB}{124,176,102}
\definecolor{DarkPink}{RGB}{255,135,135}
\definecolor{OliveGreen}{rgb}{0,0.6,0}
\DeclareRobustCommand{\hlpink}[1]{{\sethlcolor{pink}\hl{#1}}}
\DeclareRobustCommand{\hllightgreen}[1]{{\sethlcolor{LimeGreen}\hl{#1}}}

\usepackage{tikz}

\DeclareFloatingEnvironment[listname=Box,name=Box]{story}
\usepackage{adjustbox}
\usepackage{array}
\usepackage{xcolor}
\usepackage{color, colortbl}
\newcolumntype{R}[2]{%
    >{\adjustbox{angle=#1,lap=\width-(#2)}\bgroup}%
    l%
    <{\egroup}%
}
\def\basiceval#1{\the\numexpr#1\relax}
\def\cca#1{\cellcolor{blue!#10}\ifnum #1>4\color{white!#10}\else\color{black!100}\fi{.#1}}
\def\ccb#1{\cellcolor{red!#10}\ifnum #1>4\color{white!#10}\else\color{black!100}\fi{-.#1}}
\def\ccaten#1{\cellcolor{blue!#1}\ifnum #1>40\color{white!100}\else\color{black!100}\fi{
\ifnum #1>10 .#1 \else .0#1 \fi
}}
\def\ccatentwo#1#2{\cellcolor{blue!\basiceval{#2*10}}\ifnum #2>4\color{white!#1}\else\color{black!100}\fi{#1.#2}}
\def\ccbten#1{\cellcolor{red!#1}\ifnum #1>40\color{white!#1}\else\color{black!100}\fi{
\ifnum #1>10 -.#1 \else -.0#1 \fi
}}
\def\ccbtentwo#1#2{\cellcolor{red!\basiceval{#2*5}}\ifnum #2>4\color{white!#1}\else\color{black!100}\fi{-#1.#2}}
\def\ccd#1{\cellcolor{blue!\basiceval{#1 * 2}}\ifnum #1>20\color{white!#1}\else\color{black!100}\fi{
\ifnum #1>10 .#1 \else .0#1 \fi
}}
\def\ccy#1{\cellcolor{yellow!\basiceval{#1 * 3}}\ifnum #1>50\color{white!#1}\else\color{black!100}\fi{
\ifnum #1>9 .#1 \else .0#1 \fi
}}
\definecolor{darkgreen}{HTML}{38761d}
\definecolor{darkyellow}{HTML}{ffa500}

\usepackage{microtype}

\newlength{\enumerateparindent} 

\title{\textit{NewsEdits}: A News Article Revision Dataset and a Document-Level Reasoning Challenge}

\author{Alexander Spangher$^1$, Xiang Ren$^1$,  Jonathan May$^1$, Nanyun Peng$^2$ 
\\ 
$^1$Information Sciences Institute, University of Southern California \\
  ~$^2$Computer Science Department, University of California, Los Angeles  \\ 
  \texttt{\{spangher, xiangren, jonmay\}@usc.edu, violetpeng@cs.ucla.edu}
\\}

\begin{document}
\setlength{\enumerateparindent}{\parindent}

\maketitle
\begin{abstract}
News article revision histories 
provide clues to narrative and factual evolution in news articles. To facilitate analysis of this evolution, we present the first publicly available dataset of news revision histories, \textit{NewsEdits}. 
Our dataset is large-scale and multilingual; it contains 1.2 million articles with 4.6 million versions from over 22 English- and French-language newspaper sources based in three countries, spanning 15 years of coverage (2006-2021)\footnote{We release the dataset and all code used in modeling and evaluation: \url{https://github.com/isi-nlp/NewsEdits.git}}.

We define article-level edit actions: \textit{Addition}, \textit{Deletion}, \textit{Edit} and \textit{Refactor}, and develop a high-accuracy extraction algorithm to identify these actions. To underscore the factual nature of many edit actions, we conduct analyses showing that added and deleted sentences are more likely to contain updating events, main content and quotes than unchanged sentences. 

Finally, to explore whether edit actions are predictable, we introduce three novel tasks aimed at predicting actions performed during version updates. We show that these tasks are possible for expert humans but are challenging for large NLP models. 
We hope this can spur research in narrative framing and help provide predictive tools for journalists chasing breaking news.

\end{abstract}

\section{Introduction}
\label{sct:intro}

Revision histories gathered from various natural language domains like Wikipedia \cite{grundkiewicz2014wiked}, Wikihow \cite{faruqui2018wikiatomicedits} and student learner essays \cite{zhang2015annotation} have primarily been studied to explore stylistic changes, such as grammatical error correction \cite{shah2020automatic} and argumentation design \cite{afrin2020annotation}. 
However, deeper questions about content updates and narrative evolution are underexplored: \textit{Which facts are uncertain and likely to be changed? Which events are likely to update? 
What voices and perspectives are needed to complete a narrative?}

Existing edits corpora do not address these questions due to the nature of previously studied domains: as shown in \newcite{yang2017identifying}, the distribution of edits in other domains, like Wikipedia, tend to focus on syntax or style edits. In this work, we introduce a novel domain for revision histories, \textit{news article} revision histories which, we show, covers the \textit{updating} of events. Many edits in news either (1) incorporate new information (2) update events or (3) broaden perspectives (Section \ref{sct:eda}).

\begin{figure}[t]
    \centering
    \includegraphics[width=.8\linewidth]{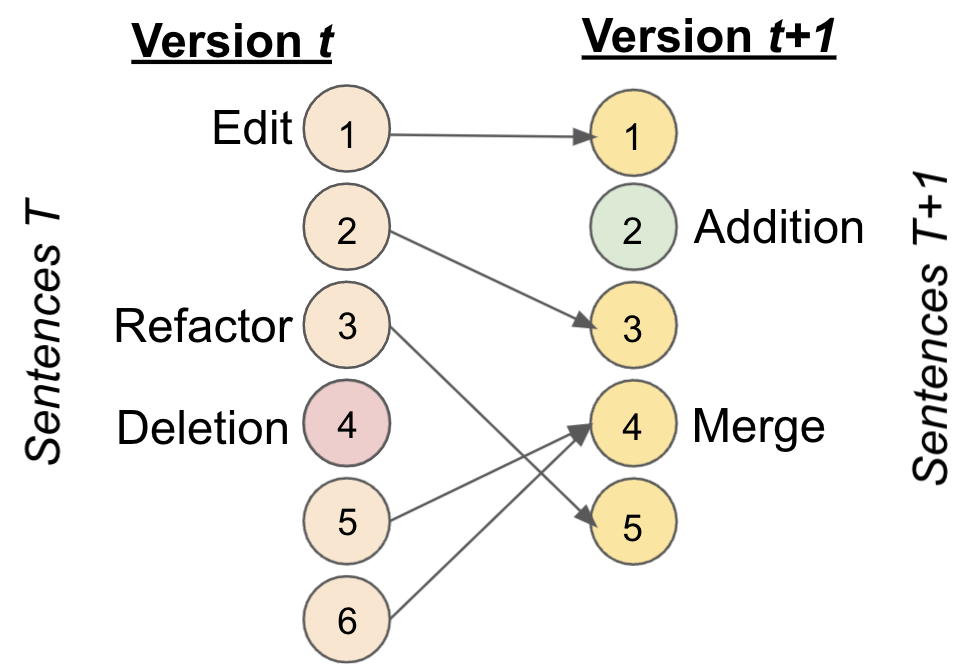}
    \caption{We identify sentence-level operations -- \textit{Edit}, \textit{Addition}, \textit{Deletion} and \textit{Refactor} -- between two versions of a news article (merges, shown here, and splits are a special cases of \textit{Edits}). We propose tasks aimed at predicting these operations on article versions. We characterize aspects of additions, deletions and edits.
    We hope \textit{NewsEdits} can contribute to research on narrative and factual development patterns.
    }
    \label{fig:sentence_diagram}
\end{figure}

Our dataset, \textit{NewsEdits}, contains 1.2 million articles with 4.6 million versions. We develop a document-level view for studying revisions and define four edit actions to characterize changes between versions: sentence \textit{Addition}, \textit{Deletion}, \textit{Edit} and \textit{Refactor} (i.e. the sentence is moved within a document). We introduce algorithms for identifying these actions. We count over 40 million 
\textit{Edits}, 
\textit{Additions}, 
\textit{Deletions} or 
\textit{Refactors} in \textit{NewsEdits}.

We argue that \textit{news is an important, practical medium to study questions about narrative, factual and stylistic development}. This is because, we hypothesize, there are consistent patterns in the way articles update in the breaking news cycle \cite{usher2018breaking}
. To prove this hypothesis, we show that updates are predictable. We design three tasks: (1) ``predict whether an article will be updated,'' (2) ``predict how much of an article will updated,'' (3) ``predict sentence-level edit actions.'' We show that current large language model (LLM)-based predictors provide a strong baseline above random guessing in most tasks, though expert human journalists perform significantly better. Our insights are twofold: (a) article updates are predictable and follow common patterns which humans are able to discern (b) significant modeling progress is needed to address the questions outlined above. See Section \ref{sct:evaluators} for more details.  

Finally, we show that the \textit{NewsEdits} dataset can bring value to a number of specific, ongoing research directions: event-temporal relation extraction \cite{ning2018multi,han2019deep}, article link prediction \cite{shahaf2010connecting}, fact-guided updates \cite{shah2020automatic}, misinformation detection \cite{appelman2015news}, headline generation \cite{shen2017recent} and author attribution \cite{savoy2013authorship}, as well as numerous directions in computational journalism \cite{cohen2011computational, spangher2020sourcefinding} and communications fields \cite{spangher2021annenberg}. 


\begin{mdframed}
Our contributions are the following:

1. We introduce \textit{NewsEdits}, the first public academic corpus of news revision histories.

2. We develop a document-level view of structural edits and introduce a highly scalable sentence-matching algorithm to label sentences in our dataset as \textit{Addition}, \textit{Deletion}, \textit{Edit}, \textit{Refactor}. We use these labels to conduct analyses characterizing these operations.

3. We introduce three novel prediction tasks to assess reasoning about whether and how an article will change. We show that current large language models perform poorly compared with expert human judgement.
\end{mdframed}



\section{The \textit{NewsEdits} Dataset}
\label{sct:dataset}

\begin{figure}
\vspace{-0.3cm}
    \centering
    \includegraphics[width=.8\linewidth]{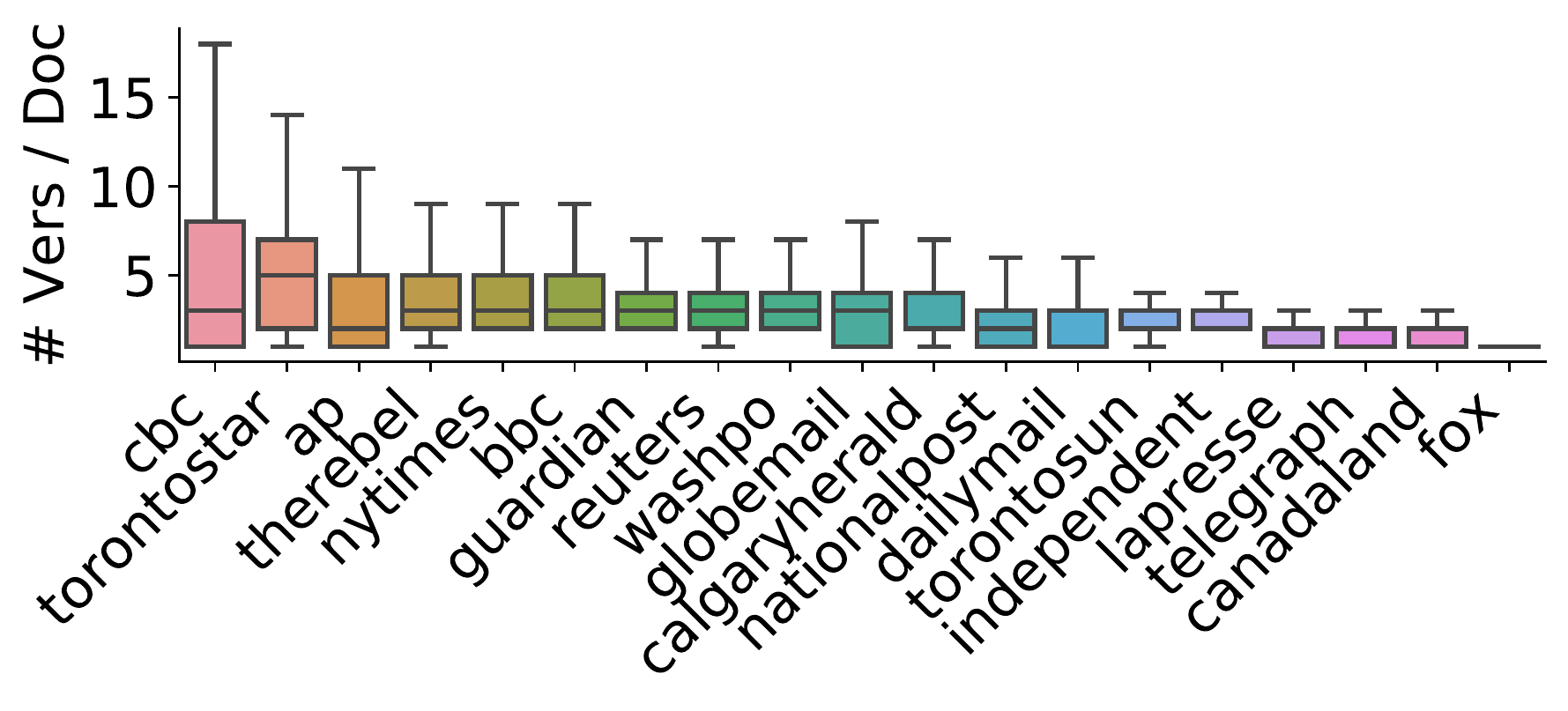}
    \vspace{-0.1cm}
    \caption{Number of versions per article, by outlet.}
    \label{fig:versions_per_article}
\vspace{-0.3cm}
\end{figure}


\textit{NewsEdits} is a dataset of 1.2 million articles and 4.6 million versions
. In Section \ref{sct:dataset:data_collection}, we discuss the sources from which we gathered our dataset. In Section \ref{sct:dataset:sentence_op_defs}, we discuss the categories of edit-actions designed to characterize changes between versions, and in Section \ref{sct:dataset:matching}, we discuss the algorithm we built to identify these edit-actions.

\subsection{Data Collection}
\label{sct:dataset:data_collection}
We collect a dataset of news article versions. An article is defined by a unique URL, while a version is one publication (of many) to that same URL. We combine data from two online sources that monitor news article updates: NewsSniffer\footnote{\url{https://www.newssniffer.co.uk/}} and Twitter accounts powered by DiffEngine\footnote{\url{https://github.com/DocNow/diffengine}}. These sources were chosen because, together, they tracked most major U.S., British and Canadian news outlets \cite{kirchhoff2010us}. Our corpus consists of article versions from 22 media outlets over a 15-year timescale (2006-2021), including \textit{The New York Times}, \textit{Washington Post} and \textit{Associated Press}. Although the median number of updates per article is 2, as shown in Figure \ref{fig:versions_per_article}, this varies depending on the outlet. More dataset details
in Appendix \ref{app:dataset_details}.

\begin{table*}[t]
    \centering
    \small
    \begin{tabular}{llrlrlr}
        \toprule
        \multicolumn{3}{c}{BERT-Based} & \multicolumn{2}{c}{Subsequence Matching} & \multicolumn{2}{c}{BLEU-Based} \\
        \cmidrule(lr){1-3}
        \cmidrule(lr){4-5}
        \cmidrule(lr){6-7}
        \multicolumn{2}{c}{ Method} & F1-Score & Method & F1-Score & Method & F1-Score \\
         \midrule
        \multirow{3}{*}{Hungarian} & TB-mini  & 88.5 &  ngram-1 & 86.0 & BLEU-1 & 86.7 \\
        & TB-medium & 88.7 & ngram-2 & 88.7 & BLEU-2 & 89.2 \\
        & RB-base & 88.6 & ngram-3 & 88.5 & BLEU-3 & 88.8 \\
        \multirow{3}{*}{Max}& TB-mini & 89.0 & ngram-4 & 88.2 & BLEU-1,2 & 88.8\\
        & TB-medium & \textbf{89.5} &&& BLEU-1,2,3 &     89.1\\
        & RB-base & 89.4 &&&&\\
        \bottomrule
    \end{tabular}
    \caption{F1 scores on validation data for matching algorithms. Left-hand group shows embedding-based methods (TinyBert (TB) and RoBERTa (RB)) with Maximum or Hungarian matching. Middle group shows ngram methods. Right-hand group shows BLEU for different ngram weightings (1,2 and 1,2,3 are uniform weightings over unigrams, bigrams and trigrams).}
    \label{tab:matching_results}
\end{table*}

\subsection{Edit-Action Operations}
\label{sct:dataset:sentence_op_defs}
Since we are interested in how an entire news article updates between versions, we focus on sentence edits (document-level actions), not word edits (sentence-level actions). Identifying that sentences are added and deleted (vs. updated), can help us study the degree of change an edit introduces in the article 
\cite{daxenberger2012corpus, daxenberger2013automatically, fong2010did}. 

Thus, we define the following sentence-level edit-actions, shown in Figure \ref{fig:sentence_diagram}: \textit{Addition}, \textit{Deletion}, \textit{Edit} and \textit{Refactor}. \textit{Additions} should contain novel information and \textit{Deletions} should remove information from the article. \textit{Edits} should be substantially similar except for syntactic changes, rephrased and minimally changed or updated information. Special cases of the \textit{Edit} operation result in sentences that are merged or split without substantial changes. See Section \ref{sct:dataset:matching} for more details.

\textit{Refactors} are intentionally moved in an article\footnote{As an example, in Figure \ref{fig:sentence_diagram}, the addition of Sentence 2 in version$_{t+1}$ shifts Sentences 3, 4, 5 down. These are \textit{not} refactors, just incidental moves caused by other operations. However, Sentences 5, 6 in version$_t$ are shifted upwards in version$_{t+1}$, which is movement that is not caused by other operations. We label this as a \textit{Refactor}.}. \textit{Refactors} are important because, based on the \textit{inverse pyramid}\footnote{ An inverse pyramid narrative structure is when the most crucial information, or purpose of the story, is presented first \cite{scanlan2003writing}.} \cite{po2003news} of article structure, sentences that are higher in an article are more important \cite{scanlan2003writing}. Thus, \textit{Refactors} give us insight into the changing importance of sentences in a narrative. 


\subsection{Edit-Action Extraction}
\label{sct:dataset:matching}
To extract these edit-actions, we need to be able to construct a bipartite graph linking sentences between two versions of an article (example graph shown in Figure \ref{fig:sentence_diagram}). If an edge exists between a sentence in one version and a sentence in the other, the sentence is an \textit{Edit} (or \textit{Unchanged}). If no edge exists, the sentence is an  \textit{Addition} (if the sentence exists in the newer version only) or \textit{Deletion} (if it exists in the older version only). We identify \textit{Refactors} based on an algorithm we develop: in short, we identify a minimal set of edges in the graph which causes all observed edge-crossings. For details on this algorithm, see Appendix \ref{app:alg_details}.

In order to construct this bipartite graph, we need a scalable, effective, sentence-similarity algorithm. There is a wide body of research in assessing sentence-similarity \cite{quan2019efficient, abujar2019sentence, yao2018novel, chen2018sentence}. 
However, many of these algorithms measure \textit{symmetric} sentence-similarity. As shown in Figure \ref{fig:sentence_diagram}, two sentences from the old version can be merged in the new version\footnote{E.g. ``ipsum\hlpink{.} Lorem'' $\rightarrow$ ``ipsum\hllightgreen{; and} Lorem''. Conversly, one sentence can also be split.}. The symmetric similarity between these three sentences would be low, leading us to label the old sentences as \textit{Deletions} and the new one an \textit{Addition}, even if they were minimally edited (for concrete examples, see Table \ref{tbl:demos}). This violates our tag definitions (Section \ref{sct:dataset:sentence_op_defs}). So, we need to measure one-way similarity between sentences, allowing us to label merged and split sentences as \textit{Edits}. Our algorithm is an asymmetrical version of the \textit{maximum alignment} metric described by \newcite{kajiwara2016building}:


\begin{align*}
\text{Sim}_{asym}(x, y) &= \frac{1}{|x|}\sum_{i=1}^{|x|}\max_j \phi(x_i, y_j)
\end{align*}

\noindent where $\phi(x_i, y_j) :=$ similarity between words $x_i$ in sentence $x$ and $y_j$ in sentence $y$.

We test several word-similarity functions, $\phi$. The first uses a simple lexical overlap, where $\phi(x_i, y_j) = 1$ if $lemma(x_i) = lemma(y_j)$ and 0 otherwise\footnote{We extend this to non-overlapping ngram matches.}. The second uses word-embeddings, where $\phi(x_i, y_j) = Emb(x_i) \cdot Emb(y_j)$, and $Emb(x_i)$ is the embedding derived from a pretrained language model \cite{jiao2019tinybert,liu2019roberta}. 

Each $\phi$ function assesses word-similarity; the next two methods use $\phi$ to assess sentence similarity. \textit{Maximum alignment} counts the number of word-matches between two sentences, allowing many-to-many word-matches between sentences. 
Hungarian matching \cite{kuhn1955hungarian} is similar, except it only allows one-to-one matches. We compare these with BLEU variations \cite{papineni2002bleu}, which have been used previously to assess sentence similarity 
\cite{faruqui2018wikiatomicedits}. 

\subsection{Edit-Action Extraction Quality}
Although our sentence-similarity algorithm is unsupervised, we need to collect ground-truth data in order to set hyperparameters (i.e. the similarity threshold above which sentences are considered a match) and evaluate different algorithms. To do this, we manually identify sentence matches in 280 documents. We asked two expert annotators to identify matches if sentences are nearly the same, they contain the same information but are stylistically different, or if they have substantial overlap in meaning and narrative function. See Appendix \ref{app:annotation} for more details on the annotation task. We use 50\% of these human-annotated labels to set hyperparameters, and 50\% to evaluate match predictions, shown in Table \ref{tab:matching_results}. Maximum Alignment with TinyBERT-medium embeddings \cite{jiao2019tinybert} (\textbf{Max-TB-medium}) performs best\footnote{For more details and examples, see Appendix \ref{app:alg_details}.}.

\begin{table}
    \centering
    \small 
    \begin{tabular}{lrr}
    \toprule
    {} &  Total Num. &  \% of Sents. \\
    \midrule
    Edits     &        26.6 mil. &         17.6 \% \\
    Additions &        10.2 mil. &          6.8 \% \\
    Deletions &         5.4 mil. &          3.6 \% \\
    Refactors &         1.6 mil. &          1.1 \% \\
    \bottomrule
    \end{tabular}
    \caption{Summary Statistics for Sentence Operations}
    \label{tbl:sum_stats}
\end{table}
 
\section{Exploratory Analysis}
\label{sct:eda}
We extract all edit actions in our dataset using methods described in the previous section. Statistics on the total number of operations are shown in Table \ref{tbl:sum_stats}.  In this section, we analyze \textit{Additions}, \textit{Deletions} and \textit{Edits} to explore when, how and why these edit-actions are made and the clues this provides as to why articles are updated. We leave a descriptive analysis of \textit{Refactors} to future work. 

\begin{figure}[t]
\centering
\subfloat[Edit-actions by version (\% total in version).]{%
  \includegraphics[width=.45\linewidth]{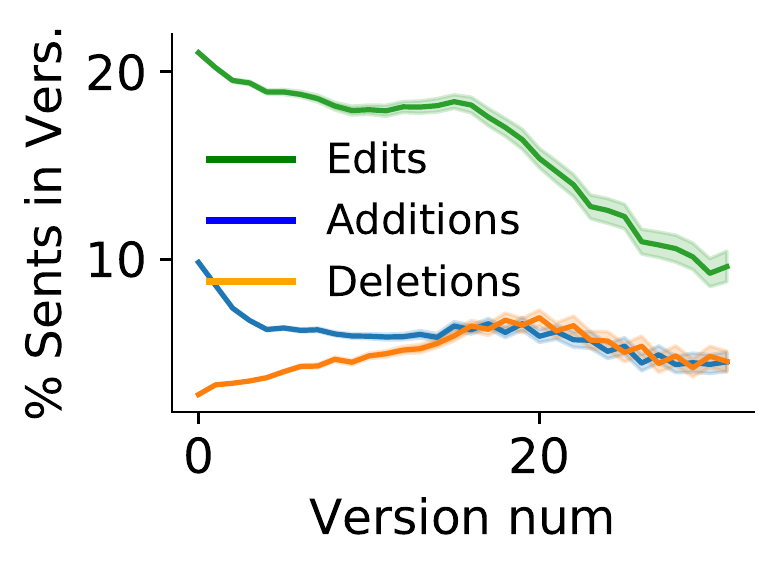}%
  \label{fig:article_dynamics:changed_sents}%
} \hspace{.05cm}
\subfloat[Edit-actions by sentence position (\%).]{
    \includegraphics[width=.45\linewidth]{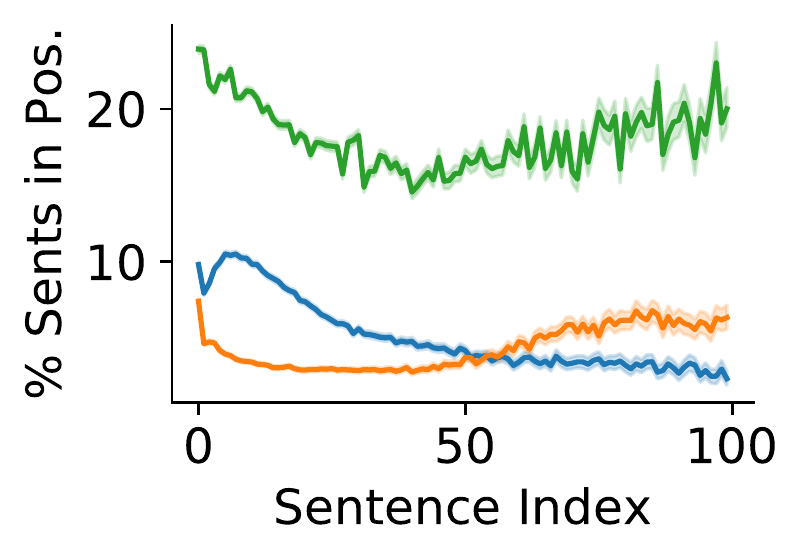}
    \label{fig:article_dynamics:pos_change}
}
\caption{Dynamics of edit actions.}
\end{figure}

\paragraph{Insight \#1: Timing and location of additions, deletions and edits reflect patterns of breaking news and inverse pyramid article structure.}
How do editing operations evolve from earlier to later versions, and where do they occur in the news article?

In Figure \ref{fig:article_dynamics:changed_sents}, we show that edit-actions in an article's early versions are primarily adding or updating information: new articles tend to have roughly 20\% of their sentences edited, 10\% added and few deleted. This fits a pattern of breaking news lifecycles: an event occurs, reporters publish a short draft quickly, and then they update as new information is learned \cite{hansen1994local, lewis2009thirst}. We further observe, as is demonstrated in Figure \ref{fig:update_time} in the appendix, that updates occur rapidly: outlets known for breaking news\footnote{E.g. \textit{Associated Press}, \textit{New York Times} and \textit{Wash. Post}} have a median article-update time of $<2$ hours.

An article's later lifecycle, we see, is determined by churn: $\approx 5\%$ of sentences are added and $5\%$ are deleted every version. As seen in Figure \ref{fig:article_dynamics:pos_change}, additions and edits are more likely to occur in the beginning of an article, while deletions are more likely at the end, indicating newer information is prioritized in an inverse pyramid structural fashion.

\begin{table}[t]
    \centering
    \small
    \begin{tabular}{lrrr}
        \addlinespace[-\aboverulesep] 
    \toprule
    \multicolumn{1}{c}{} & Add. & Del. & Unchang. \\
    \midrule
    Contains Event & 38.5 & 39.3 & 31.4 \\
    Contains Quote & 48.4 & 50.0 & 39.2 \\
    \midrule
    Discourse: Main & 4.4 & 4.9 & 3.6 \\
    Discourse: Cause & 29.0 & 30.2 & 23.6 \\
    Discourse: Distant & 63.5 & 61.4 & 68.1\\
    \bottomrule
    \end{tabular}
    \caption{\% \textit{Addition}s, \textit{Deletion}s or Unchanged sentences that contain Events or Quotes, or have news discourse role: Main (main events), Cause (immediate context) or Distant (history, analysis). $F<.01$, $n=7,368,634$.}
    \label{tab:add_sentence_attr}
\end{table}

\paragraph{Insight \#2: Additions and deletions are more likely to contain fact-patterns associated with breaking news (quotes, events, or main ideas) than unchanged sentences.}

In the previous section, we showed that the timing and position of edit-actions reflects breaking news scenarios. To provide further clues about the semantics of edit-actions, we sample \textit{Additions}, \textit{Deletions} and unchanged sentences and study the kinds of information contained in these sentences. We study three different fact-patterns associated with breaking news: events, quotes and main ideas \cite{ekstrom2021epistemologies, usher2018breaking}. To measure the prevalence of these fact-patterns, we sample 200,000 documents (7 million sentences) from our corpus and run an event-extraction pipeline \cite{ma2021eventplus}, quote-detection pipeline \cite{spangher2020sourcefinding}, and news discourse model \cite{spangher2021multitask}. As shown in Table \ref{tab:add_sentence_attr}, we find added and deleted sentences have significantly more events, quotes and \textit{Main-Idea} and \textit{Cause} discourse than unchanged sentences. (See Appendix \ref{app:eda_details} for more details.)

\paragraph{Insight \#3: Edited sentences often contain updating events.}
The analyses in the previous sections have established that edit-actions both are positioned in the article in ways that resemble, and contain information that is described by, breaking news epistemologies \cite{ekstrom2021epistemologies}. A remaining question is whether the edit-actions change fact-patterns themselves, rather than simply changing the style or other attributes of sentences.

One way to measure this is to explore whether edit-actions update the events in a story \cite{han2019joint}. We focus on pairs of edited sentences. We randomly sample \textit{Edits} from documents in our corpus ($n=432,329$ pairs) and extract events using \newcite{ma2021eventplus}'s model. We find that edited sentence pairs are more likely to contain events (43.5\%) than unchanged sentences (31.4\%). Further, we find that
37.1\% of edited sentences with events contain \textit{different} across versions. We give a sample of pairs in Table \ref{tab:event_updates}. This shows that many \textit{within} sentence operations update events.

Taken together, we have shown in this analysis that \textit{factual} updates drive many of the edit operations that we have constructed to describe \textit{NewsEdits} revision histories. Next, we will measure how predictable these update patterns are.

\begin{table}[t]
    \centering
    \small
        \begin{tabular}{p{7.5cm}}
        \toprule
        Event Chains \\
        \midrule
        (attack, killed),
        (injured, killed),
        (shot, dead),
        (shot, killed),
        (attack, injured),
        (injured, died),
        (election, won),
        (meeting, talks),
        (talks, meeting), 
        (elections, election),
        (war, conflict)\\
        \bottomrule
        \end{tabular}
    \caption{Selection of top event extracted from edited sentence pairs across article versions.}
    \label{tab:event_updates}
\end{table}

\section{Predictive Analysis on NewsEdits}
\label{sct:tasks}

As shown in Section \ref{sct:eda}, many edit-actions show breaking news patterns, which \newcite{usher2018breaking} observed follow common update patterns. Now, we explore how predictable these operations are, to address whether future work on the fundamental research questions addressed in Section \ref{sct:intro} around narrative design is feasible.

In this section, we outline three tasks 
that involve predicting the future states of articles based on the current state. These tasks, we hypothesize, outline several modeling challenges: (1) identify indicators of uncertainty used in news writing\footnote{E.g. ``Police to release details of the investigation.''} \cite{ekstrom2021epistemologies}, (2) identify informational incompleteness, like source representation \cite{spangher2020sourcefinding} and (3) identify prototypical event patterns \cite{wu2022procedural}. These are all strategies that expert human evaluators used when performing our tasks (Section \ref{sct:evaluators}). The tasks range from easier to harder, based on the sparsity of the data available for each task and the dimensionality of the prediction. We show that they are predictable but present a challenge for current language modeling approaches: expert humans perform these tasks much more accurately than LLM-based baselines.

In addition to serving a model-probing and data-explanatory purpose, these tasks are also practical: journalists told us in interviews that being able to perform these predictive tasks could help newsrooms allocate reporting resources in a breaking news scenario\footnote{See Appendix \ref{app:broader_scope} for more details.}.


\subsection{Task Description and Training Data Construction}
\label{sct:tasks:dataset_construction}

We now describe our tasks. For all three tasks, we focus on breaking news by filtering \textit{NewsEdits} down to short articles (\# sents $\in$ [5, 15]) with low version number (<20) from select outlets\footnote{The \textit{New York Times}, \textit{Associated Press}, \textit{Washington Post}, \textit{BBC}, \textit{Independent}, \textit{Guardian} and \textit{Reuters} were used, as they are more known for breaking news \cite{usher2018breaking}. See Appendix \ref{app:dataset_details} for more details.}.

\noindent\textbf{Task 1: Will this document update?} Given the text of an article at version $v$, predict if $\exists v+1$. This probes whether the model can learn a high-level notion of change, irrespective of the fact that different edit-actions have different consequences for the information presented in a news article.

For \textbf{Task 1}, $y=1$ if a newer version of an article was published and $0$ otherwise. We sample $100,000$ short article versions from \textit{NewsEdits}, balancing across length, version number, and $y$. 

\noindent\textbf{Task 2: How much will it update?} Given the text of an article at version $v$, predict in the next version how many \textit{Additions}, \textit{Deletions}, \textit{Edits}, \textit{Refactors} will occur. This moves beyond Task \#1 and requires the model to learn more about \textit{how} each edit-action category changes an article. 

For \textbf{Task 2}, $y=$ counts of sentence-level labels (\textit{Num. Additions}, \textit{Num. Deletions}, \textit{Num. Refactors}, \textit{Num. Edits}) described in the previous sections, aggregated per document. Each count is binned: $[0, 1)$, $[0, 3)$, $[3, \infty)$ and is predicted separately as a multiclass classification problem. We sample $150,000$ short article versions balancing for sources, length and version number.

\noindent\textbf{Task 3: How will it update?} For each sentence in version $v$, predict whether: (1) the sentence itself will change (i.e. it will be a \textit{Deletion} or \textit{Edit}) (2) a \textit{Refactor} will occur (i.e. it will be moved either up or down in the document) or (3) an \textit{Addition} will occur (i.e. either above or below the sentence). This task, which we hypothesize is the hardest task, requires the model to reason specifically about the informational components of each sentence \textit{and} understand nuance about structure and form in a news article (i.e. like the inverse pyramid structure \cite{po2003news}).

 For \textbf{Task 3}, $y=$ individual sentence-level labels. Labels are derived for the following subtasks mentioned above: (1) \textit{Sentence Operations} is a categorical label comprising: [Deletion, Edit, Unchanged], expressed as a one-hot vector. (2) \textit{Refactor} is a categorical label comprising: [Up, Down, Unchanged], also expressed as a one-hot vector. (3) \textit{Addition Above} and \textit{Addition Below} are each binary labels expressing whether $>1$ sentences was added above or below the target sentence. Because some sentences had \textit{Additions} above and below, we chose to model this subtask as two separate classification tasks. We sample $100,000$ short article versions, balancing for sources, length and version number.

For each task, the input $X$ is a document represented as a sequence of sentences. For each evaluation set, we sample $4k$ documents balancing for class labels (some labels are highly imbalanced and cannot be balanced).

\begin{figure}[t]
    \centering
    \includegraphics[width=.7\linewidth]{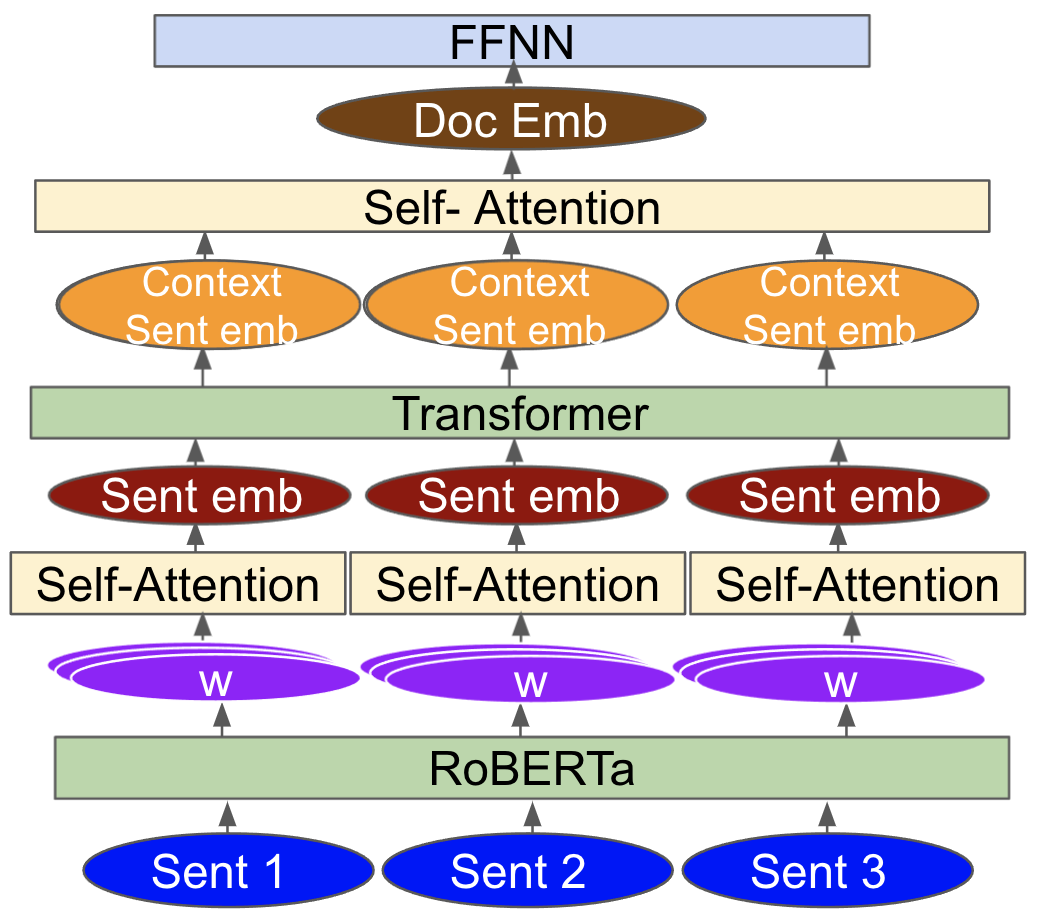}
    \caption{Architecture diagram for the model used for our tasks. Word-embeddings are averaged using \textbf{Self-Attention} to form sentence-vectors. A minimal transformer layer is used to contextualize these vectors (+Contextual Layer). In Tasks 1 and 2, self-attention is used to generate a document-embedding vector.}
    \label{fig:model_diagram}
\end{figure}

\begin{table*}[t]
\centering
\small
\begin{tabular}{lrrrrrrrr}
\toprule
 & \multicolumn{2}{c}{Num. Additions} & \multicolumn{2}{c}{Num. Deletions} & \multicolumn{2}{c}{Num. Edits} & \multicolumn{2}{c}{Num. Refactors} \\
\cmidrule(lr){2-3}\cmidrule(lr){4-5}\cmidrule(lr){6-7}\cmidrule(lr){8-9}
 &  Mac F1 &  Mic F1 &  Mac F1 & Mic F1 &  Mac F1 &  Mic F1 &  Mac F1 & Mic F1 \\
\midrule
Most Popular &           19.8 &              25.0 &               25.6 &                  47.8 &             21.9 &                32.0 &                39.2 &                   64.5 \\
Random     &             32.5 &              33.9 &               30.2 &                  36.4 &             31.7 &                35.1 &                25.8 &                   35.1 \\
\midrule
Baseline ($n=30,000$) &  22.1 & 27.9 & 25.6 & 46.5 & 21.4 & 30.6 & 35.2 & 64.5 \\
($n=150,000$) &          29.7 &              36.3 &               25.7 &                  48.1 &             22.4 &                32.8 &                39.2 &                   64.6 \\
+Partially Frozen  & \textbf{52.2} &         54.0 &               44.8 &                  59.0 &             49.3 &                53.1 &                44.3 &           \textbf{65.6} \\
+Contextual  &           50.7 &              52.2 &               41.0 &                  57.4 &             \textbf{50.8} &       \textbf{54.8} &       \textbf{45.0} &          64.3 \\
+Version     &           52.0 &      \textbf{54.5} &      \textbf{45.3} &                 \textbf{59.8} &    49.9 &                53.7 &                43.8 &                   63.1 \\
+Multitask   &           46.7 &              50.2 &               28.2 &                  48.4 &             42.1 &                49.5 &                40.3 &                   55.1 \\
\midrule
\midrule
Human        &           \textbf{66.4} &              \textbf{69.3} &               \textbf{64.6} &                  \textbf{67.5} &             \textbf{65.9} &                \textbf{75.6} &                \textbf{71.3} &                   \textbf{70.7} \\
\bottomrule
\end{tabular}
\caption{\textbf{Task 2 Benchmarks:} Baseline model performance for document-level update tasks. Counts of Added, Deleted, Edited and Refactored sentences are binned into roughly equal-sized ``low'' ($[0, 1)$ sentences), ``medium'' ($[0, 3)$ sentences), ``high'' ($[3, \infty)$ sentences) bins. Macro and Micro F1 calculated across bins. (Scores shown are median of $1,000$ bootstrap resamples of the evaluation dataset.)}
\label{tab:document-level-results}
\end{table*}

\begin{table*}[t]
\small
\centering
    \begin{tabular}{lllllll}
\toprule
{} & \multicolumn{2}{c}{Additions} &  \multicolumn{2}{c}{Sentence Operations} & \multicolumn{2}{c}{Refactors}\\
{} & Above (F1) & Below (F1)                  & Mac. F1 & Mic. F1 &                  Mac. F1 & Mic. F1 \\
\midrule
Most Popular &                 0.0 &            0.00 &                    18.1 &                       20.2 &        34.7    &   53.3  \\
Random     &          \textbf{11.8} &   \textbf{14.4} &                   28.0 &                       38.3 &        24.7    &   34.7  \\
\midrule
Baseline     &                 8.3 &            0.1 &                    \textbf{36.5} &      \textbf{61.9} &        35.2    &   54.2  \\
+Partially Frozen &          3.5 &            0.0 &                    35.4 &                        60.9 &        35.4    &   54.6  \\
+Version            &          0.1 &            0.0 &                    30.3 &                        59.0 &        \textbf{41.6}    &   \textbf{57.2}  \\
+Multitask.         &          0.0 &            0.0 &                    27.5  &                       57.8 &        39.5    &   54.8     \\
\midrule
\midrule
Human         &             \textbf{38.6} &                 \textbf{46.7} &                     \textbf{63.8} &                        \textbf{63.5} &                     \textbf{45.6} &                        \textbf{91.5} \\
\bottomrule
\end{tabular}
\caption{\textbf{Task 3 Benchmarks:} Baseline model performance for sentence-Level tasks. \textit{Addition} tasks are: ``Was a sentence added \textit{below} the target sentence?'', ``Was a sentence added \textit{above} the target sentence?'' \textit{Sentence Operations} columns are three operations that occur on the target sentence: ``Deletion'', ``Editing'', ``Unchanged''. \textit{Refactor} is binned into whether the target sentence is ``Moved Up'', ``Moved Down'' or ``Unchanged''. (Scores shown are median of $1,000$ bootstrap resamples of the evaluation dataset.)}
\label{tab:sentence-level-results}
\end{table*}

\begin{table}[t]
    \centering
    \small 
    \begin{tabular}{lrlr}
    \toprule
        & F1 & {} & F1\\
    \cmidrule(lr){1-2}\cmidrule(lr){3-4}
    Most Popular & 56.6 &     Baseline &      60.8 \\
    Random  & 50.6      &    +Partially Frozen &      66.0 \\
    Human & \textbf{80.1}       &    +Contextual &              61.7 \\
    {} &    {}          &    +Version &              \textbf{77.6 }\\
    \bottomrule
    \end{tabular}
    \caption{\textbf{Task 1 Benchmarks:} Baseline model performance for next-version prediction task. Label is binary. (Scores are median of $1,000$ bootstrap resamples of the evaluation dataset.)}
    \label{tab:version-prediction-results}
\end{table}






\subsection{Modeling}
\label{sct:tasks:modeling}

We benchmark our tasks using a RoBERTa-based architecture shown in Figure \ref{fig:model_diagram}. \newcite{spangher2021multitask} showed that a RoBERTa-based architecture \cite{liu2019roberta} with a contextualization layer outperformed other LLM-based architectures like \newcite{reimers2019sentence} for document-level understanding tasks (further insight given in Section \ref{sct:evaluators}).

In our model, each sentence from document $d$ is fed into a pretrained \texttt{RoBERTa Base} model\footnote{We used \newcite{wolf2019huggingface}'s version, found here \url{https://huggingface.co/roberta-base}.} to obtain contextualized word embeddings. The word embeddings are then averaged using self-attention, creating sentence vectors. For \textbf{Task 3}, these vectors are then used directly for sentence-level predictions. For \textbf{Tasks 1} and \textbf{2} these vectors are condensed further, using self-attention, into a single document vector which is then used for document-level predictions. The sentence vectors are optionally contextualized to incorporate knowledge of surrounding sentences, using a small Transformer layer\footnote{Specifically, we initialize a 2-layer, 2-headed GPT2 transformer block to perform autoregressive contextualization.} (+Contextualized in Tables \ref{tab:document-level-results}, \ref{tab:sentence-level-results}, \ref{tab:version-prediction-results}). 

We experiment with the following variations. For \textbf{Task 2}, we train with less data ($n=30,000$ version pairs) and more data ($n=150,000$ version pairs), balanced as described in Section \ref{sct:tasks:dataset_construction}, to test whether a larger dataset would help the models generalize better. We also experiment, for all tasks, with freezing the bottom 6 layers of the RoBERTa architecture (+Partially Frozen) to probe whether pretrained knowledge is helpful for these tasks. Additionally, we experiment giving the version number of the older version as an additional input feature alongside the text of the document (+Version).

Finally, for \textbf{Tasks 2} and \textbf{3}, we attempt to jointly model all subtasks
using separate prediction heads for each subtask but sharing all other layers. We use uniform loss weighting between the tasks. \newcite{spangher2021multitask} showed that various document-level understanding tasks could benefit by being modeled jointly. For our tasks, we hypothesize that decisions around one operation might affect another: i.e. if a writer deletes many sentences in one draft they might also add sentences, so we test whether jointly modeling has a positive effect.

We do not consider any feature engineering on the input text, like performing event extraction \cite{ma2021eventplus}, even though results in Section \ref{sct:eda} show that certain types of edit-actions are more likely to contain events. We wish to establish a strong baseline and test whether models can learn salient features on their own. For more discussion on modeling choices and hyperparameter values, see Appendix \ref{app:exp_details}. 

\subsection{Human Performance}

To evaluate how well human editors agree on edits, we design two human evaluation tasks and recruit 5 journalists with $\geq 1$ year of editing experience at major U.S. and international media outlets. 

\noindent\textbf{Evaluation Task 1:} We show users the text of an article and ask them whether or not there will be an update. Collectively, they annotate 100 articles. After completing each round, they are shown the true labels. This evaluates \textbf{Task 1}.

\noindent\textbf{Evaluation Task 2:} We show users the sentences of an article, and they are able to move sentences, mark them as deleted or edited, and add sentence-blocks above or below sentences. They are \textbf{not} asked to write any text, only mark the high-level actions of ``I \textit{would} add a sentence,'' etc. Collectively they annotate 350 news articles. After each annotation, they see what edits \textit{actually} happened. The raw output evaluates \textbf{Task 3} and we aggregate their actions for each article to evaluate \textbf{Task 2}. They are instructed to use their expert intuition and they are interviewed afterwards on the strategies used to make these predictions. (See Appendix \ref{app:annotation} for task guidelines and interviews).

\subsection{Results}

As shown in Tables \ref{tab:document-level-results}, \ref{tab:sentence-level-results}, and  \ref{tab:version-prediction-results}, model-performance indicates that our tasks do range from easier (\textbf{Task 1}) to harder (\textbf{Task 3}). While our models show improvements above \textbf{Random}, and \textbf{Most Popular} in almost all subtasks, a notable exception is \textbf{Task 3}'s \textit{Addition} subtasks, where the models do not clearly beat \textbf{Random}. We note that this was also the most difficult subtask for human evaluators.

We observe that +Partially Frozen increases performance on \textbf{Task 2}, boosting performance in all subtasks by $\approx 10$ points. In contrast, it does not increase performance on \textbf{Task 3}, perhaps indicating that the subtasks in \textbf{Task 3} are difficult for the current LLM paradigm. Although adding version embeddings (+Version) boosts performance for \textbf{Task 1}, it does not seem to measurably increase performance for the other tasks. Finally, performing \textbf{Task 2} and \textbf{3} as multitask learning problems decreases performance for all subtasks.

In contrast, human evaluators beat model performance across tasks, most consistently in \textbf{Task 2}, with on average performance 20 F1-score points above Baseline models. On \textbf{Task 3}, human performance also is high relative to model performance. We observe that, despite \textit{Additions} in \textbf{Task 3} being the hardest task, as judged by human and model performance, humans showed a $\approx 40$ point increase above model performance. Humans are also better at correctly identifying minority classes, with a wider performance gap seen for Macro F1 scores (i.e. see \textit{Sentence Operations}, where the majority of sentences are unchanged).

\begin{table}[]
    \small
    \centering
    \begin{tabular}{crcrlr}
    \toprule
        Topic ($\uparrow$) & F1 & Topic ($\downarrow$) & F1 & $y$ (Add) & F1 \\
        \cmidrule(lr){1-2}\cmidrule(lr){3-4}\cmidrule(lr){5-6}
        U.S. Pol.  &  38.1 & Local Pol. &   66.8 & [0, 1)     &  16.2 \\
        Business  &  48.4 & War &   61.8 & [1, 5)     &  59.7 \\
        U.K. Pol.   &  50.4 & Crime  &   58.3 & [5, 100)   &   0.9\\
    \bottomrule
    \end{tabular}
    \caption{Error Analysis: LDA (first two columns): Documents belonging to some topics are easier to predict than others. By label (last column): medium-range growth is easier to predict.}
    \label{tab:lda_results}
\end{table}

\subsection{Error Analysis}
\label{sct:error_analysis}
We perform an error analysis on the \textbf{Task 2} task and find that there are several categories of edits that are easier to predict than others. We run Latent Dirichlet allocation on $40,000$ articles, shown in Table \ref{tab:lda_results}\footnote{Topic words shown in Appendix \ref{app:error_analysis}.}. We assign documents to their highest topic and find that articles covering certain news topics (like \textit{War}) update in a much more predictable pattern than others (like $Business$), with a spread of over $26$ F1-score points. Further, we find that certain edit-patterns are easier to differentiate, like articles that grow between 1-5 sentences (Table \ref{tab:lda_results}). This show us ways to select for subsets of our dataset that are more standard in their update patterns. 

The class imbalance of this dataset (Table \ref{tbl:sum_stats}) results in the \textbf{Most Popular} scoring highly. To mitigate this, we evaluate on balanced datasets. Class imbalanced training approaches \cite{li2019dice, spangher2021multitask} might be of further help. 



\subsection{Evaluator Interviews}
\label{sct:evaluators}
To better understand the process involved with successful human annotation, we conducted evaluator interviews. We noticed that evaluators first identified whether the main news event was still occurring, or if it was in the past. For the former, they tried to predict when the event would update.\footnote{The longer the timespan, the more information they predicted would be added between drafts.} For the latter, they considered discourse components to determine if an article was narratively complete and analyzed the specificity of the quotes.\footnote{E.g. Generic quotes, say a public announcement, would be updated with specific, eye-witness quotes.} They determined where to add information in the story based on structural analysis
, and stressed the importance of the inverse pyramid for \textit{informational uncertainty}: information later in an article had more uncertainty; if confirmed, it would be moved up in later versions.\footnote{One evaluator called this a \textit{``buried cause''}.
} Finally, they considered the emotional salience of events; if a sentence described an event causing harm, it would be moved up\footnote{See Appendix \ref{app:annotation} for full interviews.}.

Clearly, these tasks demand strong world-knowledge and common sense, as well and high-level discourse, structural and narrative awareness.\footnote{Evaluators told us they ``thought like the AP.'' The AP, or the \textit{Associated Press}, has a styleguide \cite{apstyle} that many outlets use to guide their writing.} Combining these different forms of reasoning, our results show, is challenging for current language models, which, for many subtasks, perform worse than guessing. \textbf{+Multitask} performance actually decreases performance for both \textbf{Task 2} and \textbf{Task 3}, indicating that these models learn features that do not generalize across subtasks. This contrasts with what our evaluators said: their decision to delete sentences often used the same reasoning as, and were dependent on, their decisions to add. 

However, we see potential for improvement in these tasks. Current LLMs have been shown to identify common arcs in story-telling \cite{boyd2020narrative}, identify event-sequences \cite{han2019joint} and reason about discourse structures \cite{spangher2021multitask,li2019scientific}. Further, for the ROCStories challenge, which presents four sentences and tasks the model with predicting the fifth \cite{mostafazadeh2017lsdsem,mostafazadeh2016corpus}, LLMs have been shown to perform scene reconstruction \cite{tian2020scene}, story planning \cite{yao2019plan,peng2018towards}, and structural common sense reasoning \cite{chen2019incorporating}. These are all aspects of reasoning that our evaluators told us they relied on. Narrative arcs in journalism are often standard and structured \cite{neiger2016understanding}, so we see potential for improvement.


\section{Related Work}

A significant contribution of this work, we feel, is the introduction of a large corpus of news edits into revision-history research and the framing of questions around sentence-level edit-actions. Despite the centrality of news writing in NLP \cite{marcus1993building, carlson2003building, pustejovsky2003timebank, ace2005}, we know of no academic corpus of news revision histories. 
Two works that analyze news edits to predict article quality \cite{tamori2017analyzing, hitomi2017proofread} do not  release their datasets.\footnote{Datasets could not be released due to copyright infringement, according to the authors in response to our inquiry.} 
WikiNews\footnote{\url{https://en.wikinews.org/wiki/Main_Page}} articles and editor-annotations have been used for document summarization \cite{bravo2012zipf}, timeline synthesis \cite{zhang2017towards, minard2016meantime}, word-identification \cite{yimam2017cwig3g2} and entity salience \cite{wu2020wn}. However, we are not aware of any work using WikiNews revision histories. 
We did not include WikiNews because its collaborative community edits differ from professional news edits.

Since at least 2006, internet activists have tracked changes made to major digital news articles \cite{bbc2006}. \url{NewsDiffs.org}, NewsSniffer 
  and DiffEngine 
 are platforms which researchers have used to study instances of gender and racial bias in article drafts,\footnote{\url{http://www.newsdiffs.org/diff/192021/192137/www.nytimes.com/2013/03/31/science/space/yvonne-brill-rocket-scientist-dies-at-88.html}}
\cite{brisbane_2012, burke_2016,jones2017using,fass2014revealing} 
shifting portrayals of social events, \cite{johnson2016effect} and lack of media transparency \cite{gourarie_2015}. These tools collect article versions from RSS feeds and the Internet Archive. Major newspapers\footnote{\url{https://twitter.com/i/lists/821699483088076802}} and thousands of government websites\footnote{\url{https://envirodatagov.org/federal-environmental-web-tracker-about-page/}} are being analyzed. We use DiffEngine and NewsSniffer to construct \textit{NewsEdits}.

\textbf{Wikihow} \cite{anthonio2020wikihowtoimprove,bhat2020towards} and \textbf{Source Code Diffs} \cite{tan2019survey,shen2019intellimerge,tsantalis2018accurate,silva2017refdiff,marrese2020learning,xu2019commit} use revision histories from domains and for purposes different than ours
. Many tasks have benefited from studying \noindent\textbf{Wikipedia Revisions}, like text simplification \cite{yatskar2010sake}, textual entailment \cite{zanzotto2010expanding}, discourse learning \cite{daxenberger2013automatically} and grammatical error correction \cite{faruqui2018wikiatomicedits}. However, most tasks focus on word-level edit operations to explore sentence-level changes. Ours focuses on sentence-level operations to explore document-level changes. 
Research in \noindent\textbf{Student Learner Essays} focuses on editing revisions made during essay-writing  \cite{leacock2010automated,wang2020erevis,zhang2020engaging,zhang2015annotation}. Researchers categorize the intention and effects of each edit 
\cite{zhang2017corpus, afrin2020annotation}, but do not try to predict edits. 

\section{Conclusion}

In this work, we have introduced the first large-scale dataset of news edits, extracted edit-actions, and shown that many were fact-based. We showed that edit-actions are predictable by experts but challenging for current LM-backed classifierss. Going forward, we will develop a schema describing the types of edits
. We are inspired by the Wikipedia Intentions schema developed by \newcite{yang2017identifying}, and are working in collaboration with journalists to further clarify the differences. This development will help to clarify the nature of these edits as well as focus further directions of inquiry.

\section{Acknowledgements}

We are grateful to Amanda Stent, Sz-Rung Shiang, Gabriel Kahn, Casey Williams, Meg Robbins, I-Hung Hsu, Mozhdeh Gheini, Jiao Sun and our anonymous reviewers for invaluable feedback. Spangher is grateful for Bloomberg for supporting this research with a PhD fellowship. May is supported by DARPA Contract FA8750-19-2-0500. 

\section{Ethical Considerations}

\subsection{Dataset}

We received permission from the original owners of the datasets, NewsSniffer and DiffEngine. Both sources are shared under strong sharing licenses. NewsSniffer is released under an \texttt{AGPL-3.0 License}\footnote{\url{https://opensource.org/licenses/AGPL-3.0}}, which is a strong ``CopyLeft'' license. DiffEngine is released under an \texttt{Attribution-NoDerivatives 4.0 International} license\footnote{\scriptsize{\url{https://creativecommons.org/licenses/by-nd/4.0/}}}.

Our use is within the bounds of intended use given in writing by the original dataset creators, and is within the scope of their licensing.

\subsection{Privacy}

We believe that there are no adverse privacy implications in this dataset. The dataset comprises news articles that were already published in the public domain with the expectation of widespread distribution. We did not engage in any concerted effort to assess whether information within the dataset was libelious, slanderous or otherwise unprotected speech. We instructed annotators to be aware that this was a possibility and to report to us if they saw anything, but we did not receive any reports. We discuss this more below.  

\subsection{Limitations and Risks}

The primary theoretical limitation in our work is that we did not include a robust non-Western language source; indeed, our only two languages were English and French. We tried to obtain sources in non-Western newspapers and reached out to a number of activists that use the DiffEngine platform to collect news outside of the Western world, including activists from Russia and Brazil. Unfortunately, we were not able to get a responses. 

Thus, this work should be viewed with that important caveat. We cannot assume \textit{a priori} that all cultures necessarily follow this approach to breaking news and indeed all of the theoretical works that we cite in justifying our directions also focus on English-language newspapers. We provide documentation in the Appendix about the language, source, timeline and size of each media outlet that we use in this dataset.

One possible risk is that some of the information contained in earlier versions of news articles was updated or removed for the express purpose that it was potentially unprotected speech: libel, slander, etc. We discussed this with the original authors of NewsSniffer and DiffEngine. During their years of operation, neither author has received any requests to take versions down. Furthermore, instances of First Amendment lawsuits where the plaintiff was successful in challenging content are rare in the U.S. We are not as familiar with the guidelines of protected speech in other countries.

Another risk we see is the misuse of this work on edits for the purpose of disparaging and denigrating media outlets. Many of these news tracker websites have been used for noble purposes (e.g. holding newspapers accountable for when they make stylistic edits or try to update without giving notice). But we live in a political environment that is often hostile to the core democracy-preserving role of the media. We focus on fact-based updates and hope that this resource is not used to unnecessarily find fault with media outlets.  

\subsection{Computational Resources}

The experiments in our paper require computational resources. All our models run on a single 30GB NVIDIA V100 GPU, along with storage and CPU capabilities provided by AWS. While our experiments do not need to leverage model or data parallelism, we still recognize that not all researchers have access to this resource level.

We use Huggingface RoBERTa-base models for our predictive tasks, and release the code of all the custom architectures that we construct at \url{https://github.com/isi-nlp/NewsEdits.git}. Our models do not exceed 300 million parameters.

\subsection{Annotators}

We recruited annotators from professional journalism networks like the NICAR listserve\footnote{\url{https://www.ire.org/training/conferences/nicar-2021/}}. All the annotators consented to annotate as part of the experiment, and were paid \$1 per task, above the highest minimum wage in the U.S. Of our five annotators, three are based in large U.S. cities, one lives in a small U.S. city and one lives in a large Brazilian city. Four annotators identify as white and one identifies as Latinx. Four annotators identify as male and one identifies as female. This data collection process is covered under a university IRB. We do not publish personal details about the annotations, and their interviews were given with consent and full awareness that they would be published in full.


\FloatBarrier
\clearpage
\newpage
\bibliographystyle{acl_natbib}
\bibliography{custom}

\begin{thebibliography}{96}
\expandafter\ifx\csname natexlab\endcsname\relax\def\natexlab#1{#1}\fi

\bibitem[{Abujar et~al.(2019)Abujar, Hasan, and Hossain}]{abujar2019sentence}
Sheikh Abujar, Mahmudul Hasan, and Syed~Akhter Hossain. 2019.
\newblock Sentence similarity estimation for text summarization using deep
  learning.
\newblock In \emph{Proceedings of the 2nd International Conference on Data
  Engineering and Communication Technology}, pages 155--164. Springer.

\bibitem[{Afrin et~al.(2020)Afrin, Wang, Litman, Matsumura, and
  Correnti}]{afrin2020annotation}
Tazin Afrin, Elaine~Lin Wang, Diane Litman, Lindsay~Clare Matsumura, and
  Richard Correnti. 2020.
\newblock Annotation and classification of evidence and reasoning revisions in
  argumentative writing.
\newblock In \emph{Proceedings of the Fifteenth Workshop on Innovative Use of
  NLP for Building Educational Applications}, pages 75--84.

\bibitem[{Anthonio et~al.(2020)Anthonio, Bhat, and
  Roth}]{anthonio2020wikihowtoimprove}
Talita Anthonio, Irshad Bhat, and Michael Roth. 2020.
\newblock \href {https://aclanthology.org/2020.lrec-1.702}
  {wiki{H}ow{T}o{I}mprove: A resource and analyses on edits in instructional
  texts}.
\newblock In \emph{Proceedings of the 12th Language Resources and Evaluation
  Conference}, pages 5721--5729, Marseille, France. European Language Resources
  Association.

\bibitem[{Appelman and Hettinga(2015)}]{appelman2015news}
Alyssa Appelman and Kirstie Hettinga. 2015.
\newblock Do news corrections affect credibility? not necessarily.
\newblock \emph{Newspaper Research Journal}, 36(4):415--425.

\bibitem[{Bhat et~al.(2020)Bhat, Anthonio, and Roth}]{bhat2020towards}
Irshad Bhat, Talita Anthonio, and Michael Roth. 2020.
\newblock \href {https://doi.org/10.18653/v1/2020.emnlp-main.675} {Towards
  modeling revision requirements in wiki{H}ow instructions}.
\newblock In \emph{Proceedings of the 2020 Conference on Empirical Methods in
  Natural Language Processing (EMNLP)}, pages 8407--8414, Online. Association
  for Computational Linguistics.

\bibitem[{Blei et~al.(2003)Blei, Ng, and Jordan}]{blei2003latent}
David~M. Blei, Andrew~Y. Ng, and Michael~I. Jordan. 2003.
\newblock Latent dirichlet allocation.
\newblock \emph{J. Mach. Learn. Res.}, 3:993–1022.

\bibitem[{Boyd et~al.(2020)Boyd, Blackburn, and Pennebaker}]{boyd2020narrative}
Ryan~L Boyd, Kate~G Blackburn, and James~W Pennebaker. 2020.
\newblock The narrative arc: Revealing core narrative structures through text
  analysis.
\newblock \emph{Science advances}, 6(32):eaba2196.

\bibitem[{Bravo-Marquez and Manriquez(2012)}]{bravo2012zipf}
Felipe Bravo-Marquez and Manuel Manriquez. 2012.
\newblock A zipf-like distant supervision approach for multi-document
  summarization using wikinews articles.
\newblock In \emph{International Symposium on String Processing and Information
  Retrieval}, pages 143--154. Springer.

\bibitem[{Brisbane(2012)}]{brisbane_2012}
Arthur~S. Brisbane. 2012.
\newblock \href
  {https://www.nytimes.com/2012/07/01/opinion/sunday/article-changes-are-shown-in-a-tool-created-by-outsiders.html}
  {Insider's view of changes, from outside}.
\newblock \emph{The New York Times}.

\bibitem[{Burke(2016)}]{burke_2016}
Austin Burke. 2016.
\newblock \href
  {https://www.vrresearch.com/blog/2016/9/1/newsdiffs-a-tool-for-tracking-changes-to-online-news-articles}
  {Newsdiffs: A tool for tracking changes to online news articles - vr research
  - public records research: Opposition research}.

\bibitem[{Carlson et~al.(2003)Carlson, Marcu, and
  Okurowski}]{carlson2003building}
Lynn Carlson, Daniel Marcu, and Mary~Ellen Okurowski. 2003.
\newblock Building a discourse-tagged corpus in the framework of rhetorical
  structure theory.
\newblock In \emph{Current and new directions in discourse and dialogue}, pages
  85--112. Springer.

\bibitem[{Chen et~al.(2019)Chen, Chen, and Yu}]{chen2019incorporating}
Jiaao Chen, Jianshu Chen, and Zhou Yu. 2019.
\newblock Incorporating structured commonsense knowledge in story completion.
\newblock In \emph{Proceedings of the AAAI Conference on Artificial
  Intelligence}, volume~33, pages 6244--6251.

\bibitem[{Chen et~al.(2018)Chen, Kim, Wilbur, and Lu}]{chen2018sentence}
Qingyu Chen, Sun Kim, W~John Wilbur, and Zhiyong Lu. 2018.
\newblock Sentence similarity measures revisited: ranking sentences in pubmed
  documents.
\newblock In \emph{Proceedings of the 2018 ACM International Conference on
  Bioinformatics, Computational Biology, and Health Informatics}, pages
  531--532.

\bibitem[{Choubey et~al.(2020)Choubey, Lee, Huang, and
  Wang}]{choubey2020discourse}
Prafulla~Kumar Choubey, Aaron Lee, Ruihong Huang, and Lu~Wang. 2020.
\newblock \href {https://doi.org/10.18653/v1/2020.acl-main.478} {Discourse as a
  function of event: Profiling discourse structure in news articles around the
  main event}.
\newblock In \emph{Proceedings of the 58th Annual Meeting of the Association
  for Computational Linguistics}, pages 5374--5386, Online. Association for
  Computational Linguistics.

\bibitem[{Cohen et~al.(2011)Cohen, Hamilton, and
  Turner}]{cohen2011computational}
Sarah Cohen, James~T Hamilton, and Fred Turner. 2011.
\newblock Computational journalism.
\newblock \emph{Communications of the ACM}, 54(10):66--71.

\bibitem[{Daxenberger and Gurevych(2012)}]{daxenberger2012corpus}
Johannes Daxenberger and Iryna Gurevych. 2012.
\newblock \href {https://www.aclweb.org/anthology/C12-1044} {A corpus-based
  study of edit categories in featured and non-featured {W}ikipedia articles}.
\newblock In \emph{Proceedings of {COLING} 2012}, pages 711--726, Mumbai,
  India. The COLING 2012 Organizing Committee.

\bibitem[{Daxenberger and Gurevych(2013)}]{daxenberger2013automatically}
Johannes Daxenberger and Iryna Gurevych. 2013.
\newblock Automatically classifying edit categories in wikipedia revisions.
\newblock In \emph{Proceedings of the 2013 Conference on Empirical Methods in
  Natural Language Processing}, pages 578--589.

\bibitem[{Ekstr{\"o}m et~al.(2021)Ekstr{\"o}m, Rams{\"a}lv, and
  Westlund}]{ekstrom2021epistemologies}
Mats Ekstr{\"o}m, Amanda Rams{\"a}lv, and Oscar Westlund. 2021.
\newblock The epistemologies of breaking news.
\newblock \emph{Journalism Studies}, 22(2):174--192.

\bibitem[{Faruqui et~al.(2018)Faruqui, Pavlick, Tenney, and
  Das}]{faruqui2018wikiatomicedits}
Manaal Faruqui, Ellie Pavlick, Ian Tenney, and Dipanjan Das. 2018.
\newblock \href {https://doi.org/10.18653/v1/D18-1028} {{W}iki{A}tomic{E}dits:
  A multilingual corpus of {W}ikipedia edits for modeling language and
  discourse}.
\newblock pages 305--315.

\bibitem[{Fass and Main(2014)}]{fass2014revealing}
John Fass and Angus Main. 2014.
\newblock Revealing the news: How online news changes without you noticing.
\newblock \emph{Digital Journalism}, 2(3):366--382.

\bibitem[{Ferrara(2017)}]{ferrara2017disinformation}
Emilio Ferrara. 2017.
\newblock Disinformation and social bot operations in the run up to the 2017
  french presidential election.
\newblock \emph{arXiv preprint arXiv:1707.00086}.

\bibitem[{Fong and Biuk-Aghai(2010)}]{fong2010did}
Peter Kin-Fong Fong and Robert~P Biuk-Aghai. 2010.
\newblock What did they do? deriving high-level edit histories in wikis.
\newblock In \emph{Proceedings of the 6th International Symposium on Wikis and
  Open Collaboration}, pages 1--10.

\bibitem[{Fu et~al.(2018)Fu, Tan, Peng, Zhao, and Yan}]{fu2018style}
Zhenxin Fu, Xiaoye Tan, Nanyun Peng, Dongyan Zhao, and Rui Yan. 2018.
\newblock Style transfer in text: Exploration and evaluation.
\newblock In \emph{Proceedings of the AAAI Conference on Artificial
  Intelligence}, volume~32.

\bibitem[{Goldstein(1953)}]{apstyle}
Norm Goldstein. 1953.
\newblock \href {"https://www.apstylebook.com/"} {\emph{The Associate Press
  Rules Regulations and General Orders}}.

\bibitem[{Gourarie(2015)}]{gourarie_2015}
Chava Gourarie. 2015.
\newblock \href {https://www.cjr.org/watchdog/newsdiffs_new_york_times.php}
  {Why 'diffing' could make news organizations more transparent}.
\newblock \emph{Columbia Journalism Review}.

\bibitem[{Grundkiewicz and Junczys-Dowmunt(2014)}]{grundkiewicz2014wiked}
Roman Grundkiewicz and Marcin Junczys-Dowmunt. 2014.
\newblock The wiked error corpus: A corpus of corrective wikipedia edits and
  its application to grammatical error correction.
\newblock In \emph{International Conference on Natural Language Processing},
  pages 478--490. Springer.

\bibitem[{Gupta and Yang(2019)}]{gupta2019predicting}
Raj~Kumar Gupta and Yinping Yang. 2019.
\newblock Predicting and understanding news social popularity with emotional
  salience features.
\newblock In \emph{Proceedings of the 27th ACM International Conference on
  Multimedia}, pages 139--147.

\bibitem[{Han et~al.(2019{\natexlab{a}})Han, Hsu, Yang, Galstyan, Weischedel,
  and Peng}]{han2019deep}
Rujun Han, I-Hung Hsu, Mu~Yang, Aram Galstyan, Ralph Weischedel, and Nanyun
  Peng. 2019{\natexlab{a}}.
\newblock Deep structured neural network for event temporal relation
  extraction.
\newblock In \emph{The 2019 SIGNLL Conference on Computational Natural Language
  Learning (CoNLL)}.

\bibitem[{Han et~al.(2019{\natexlab{b}})Han, Ning, and Peng}]{han2019joint}
Rujun Han, Qiang Ning, and Nanyun Peng. 2019{\natexlab{b}}.
\newblock \href {https://doi.org/10.18653/v1/D19-1041} {Joint event and
  temporal relation extraction with shared representations and structured
  prediction}.
\newblock In \emph{Proceedings of the 2019 Conference on Empirical Methods in
  Natural Language Processing and the 9th International Joint Conference on
  Natural Language Processing (EMNLP-IJCNLP)}, pages 434--444, Hong Kong,
  China. Association for Computational Linguistics.

\bibitem[{Hansen et~al.(1994)Hansen, Ward, Conners, and
  Neuzil}]{hansen1994local}
Kathleen~A Hansen, Jean Ward, Joan~L Conners, and Mark Neuzil. 1994.
\newblock Local breaking news: Sources, technology, and news routines.
\newblock \emph{Journalism Quarterly}, 71(3):561--572.

\bibitem[{Hashimoto et~al.(2018)Hashimoto, Guu, Oren, and
  Liang}]{hashimoto2018retrieve}
Tatsunori~B. Hashimoto, Kelvin Guu, Yonatan Oren, and Percy Liang. 2018.
\newblock A retrieve-and-edit framework for predicting structured outputs.
\newblock In \emph{Proceedings of the 32nd International Conference on Neural
  Information Processing Systems}, NIPS'18, page 10073–10083, Red Hook, NY,
  USA. Curran Associates Inc.

\bibitem[{Herrmann(2006)}]{bbc2006}
Steve Herrmann. 2006.
\newblock \href
  {https://www.bbc.co.uk/blogs/theeditors/2006/10/sniffing_out_edits.html} {The
  editors: Sniffing out edits}.
\newblock \emph{BBC}.

\bibitem[{Hitomi et~al.(2017)Hitomi, Tamori, Okazaki, and
  Inui}]{hitomi2017proofread}
Yuta Hitomi, Hideaki Tamori, Naoaki Okazaki, and Kentaro Inui. 2017.
\newblock Proofread sentence generation as multi-task learning with editing
  operation prediction.
\newblock In \emph{Proceedings of the Eighth International Joint Conference on
  Natural Language Processing (Volume 2: Short Papers)}, pages 436--441.

\bibitem[{Jiao et~al.(2020)Jiao, Yin, Shang, Jiang, Chen, Li, Wang, and
  Liu}]{jiao2019tinybert}
Xiaoqi Jiao, Yichun Yin, Lifeng Shang, Xin Jiang, Xiao Chen, Linlin Li, Fang
  Wang, and Qun Liu. 2020.
\newblock \href {https://doi.org/10.18653/v1/2020.findings-emnlp.372}
  {{T}iny{BERT}: Distilling {BERT} for natural language understanding}.
\newblock In \emph{Findings of the Association for Computational Linguistics:
  EMNLP 2020}, pages 4163--4174, Online. Association for Computational
  Linguistics.

\bibitem[{Johnson et~al.(2016)Johnson, Schreiner, and
  Agnone}]{johnson2016effect}
Erik~W Johnson, Jonathan~P Schreiner, and Jon Agnone. 2016.
\newblock The effect of new york times event coding techniques on social
  movement analyses of protest data.
\newblock In \emph{Narratives of Identity in Social Movements, Conflicts and
  Change}. Emerald Group Publishing Limited.

\bibitem[{Jones and Neubert(2017)}]{jones2017using}
Gina~M Jones and Michael Neubert. 2017.
\newblock Using {RSS} to improve web harvest results for news web sites.
\newblock \emph{Journal of Western Archives}, 8(2):3.

\bibitem[{Kajiwara and Komachi(2016)}]{kajiwara2016building}
Tomoyuki Kajiwara and Mamoru Komachi. 2016.
\newblock \href {https://aclanthology.org/C16-1109} {Building a monolingual
  parallel corpus for text simplification using sentence similarity based on
  alignment between word embeddings}.
\newblock In \emph{Proceedings of {COLING} 2016, the 26th International
  Conference on Computational Linguistics: Technical Papers}, pages 1147--1158,
  Osaka, Japan. The COLING 2016 Organizing Committee.

\bibitem[{Kirchhoff(2010)}]{kirchhoff2010us}
Suzanne~M Kirchhoff. 2010.
\newblock \emph{US newspaper industry in transition}.
\newblock DIANE Publishing.

\bibitem[{Kuhn(1955)}]{kuhn1955hungarian}
Harold~W Kuhn. 1955.
\newblock The hungarian method for the assignment problem.
\newblock \emph{Naval research logistics quarterly}, 2(1-2):83--97.

\bibitem[{Leacock et~al.(2010)Leacock, Chodorow, Gamon, and
  Tetreault}]{leacock2010automated}
Claudia Leacock, Martin Chodorow, Michael Gamon, and Joel Tetreault. 2010.
\newblock Automated grammatical error detection for language learners.
\newblock \emph{Synthesis lectures on human language technologies},
  3(1):1--134.

\bibitem[{Lewis and Cushion(2009)}]{lewis2009thirst}
Justin Lewis and Stephen Cushion. 2009.
\newblock The thirst to be first: An analysis of breaking news stories and
  their impact on the quality of 24-hour news coverage in the uk.
\newblock \emph{Journalism Practice}, 3(3):304--318.

\bibitem[{Li et~al.(2021)Li, Burns, and Peng}]{li2019scientific}
Xiangci Li, Gully Burns, and Nanyun Peng. 2021.
\newblock \href {https://doi.org/10.18653/v1/2021.eacl-main.218} {Scientific
  discourse tagging for evidence extraction}.
\newblock In \emph{Proceedings of the 16th Conference of the European Chapter
  of the Association for Computational Linguistics: Main Volume}, pages
  2550--2562, Online. Association for Computational Linguistics.

\bibitem[{Li et~al.(2020)Li, Sun, Meng, Liang, Wu, and Li}]{li2019dice}
Xiaoya Li, Xiaofei Sun, Yuxian Meng, Junjun Liang, Fei Wu, and Jiwei Li. 2020.
\newblock \href {https://doi.org/10.18653/v1/2020.acl-main.45} {Dice loss for
  data-imbalanced {NLP} tasks}.
\newblock In \emph{Proceedings of the 58th Annual Meeting of the Association
  for Computational Linguistics}, pages 465--476, Online. Association for
  Computational Linguistics.

\bibitem[{Liu et~al.(2019)Liu, Ott, Goyal, Du, Joshi, Chen, Levy, Lewis,
  Zettlemoyer, and Stoyanov}]{liu2019roberta}
Yinhan Liu, Myle Ott, Naman Goyal, Jingfei Du, Mandar Joshi, Danqi Chen, Omer
  Levy, Mike Lewis, Luke Zettlemoyer, and Veselin Stoyanov. 2019.
\newblock Roberta: A robustly optimized bert pretraining approach.
\newblock \emph{arXiv preprint arXiv:1907.11692}.

\bibitem[{Lu and Peng(2021)}]{lu2021efficient}
Sidi Lu and Nanyun Peng. 2021.
\newblock On efficient training, controllability and compositional
  generalization of insertion-based language generators.
\newblock \emph{arXiv preprint arXiv:2102.11008}.

\bibitem[{Ma et~al.(2021)Ma, Sun, Yang, Huang, Wen, Singh, Han, and
  Peng}]{ma2021eventplus}
Mingyu~Derek Ma, Jiao Sun, Mu~Yang, Kung-Hsiang Huang, Nuan Wen, Shikhar Singh,
  Rujun Han, and Nanyun Peng. 2021.
\newblock Eventplus: A temporal event understanding pipeline.
\newblock \emph{arXiv preprint arXiv:2101.04922}.

\bibitem[{Marcus et~al.(1993)Marcus, Santorini, and
  Marcinkiewicz}]{marcus1993building}
Mitchell Marcus, Beatrice Santorini, and Mary~Ann Marcinkiewicz. 1993.
\newblock Building a large annotated corpus of english: The penn treebank.

\bibitem[{Marrese-Taylor et~al.(2020)Marrese-Taylor, Loyola, Balazs, and
  Matsuo}]{marrese2020learning}
Edison Marrese-Taylor, Pablo Loyola, Jorge~A Balazs, and Yutaka Matsuo. 2020.
\newblock Learning to describe editing activities in collaborative
  environments: A case study on github and wikipedia.
\newblock In \emph{Proceedings of the 34th Pacific Asia Conference on Language,
  Information and Computation}, pages 188--198.

\bibitem[{Mehrabi et~al.(2020)Mehrabi, Gowda, Morstatter, Peng, and
  Galstyan}]{mehrabi2020man}
Ninareh Mehrabi, Thamme Gowda, Fred Morstatter, Nanyun Peng, and Aram Galstyan.
  2020.
\newblock Man is to person as woman is to location: Measuring gender bias in
  named entity recognition.
\newblock In \emph{Proceedings of the 31st ACM Conference on Hypertext and
  Social Media}, pages 231--232.

\bibitem[{Minard et~al.(2016)Minard, Speranza, Urizar, Altuna, Van~Erp, Schoen,
  and Van~Son}]{minard2016meantime}
Anne-Lyse Minard, Manuela Speranza, Ruben Urizar, Begona Altuna, Marieke
  Van~Erp, Anneleen Schoen, and Chantal Van~Son. 2016.
\newblock Meantime, the newsreader multilingual event and time corpus.
\newblock In \emph{Proceedings of the Tenth International Conference on
  Language Resources and Evaluation (LREC'16)}, pages 4417--4422.

\bibitem[{Mostafazadeh et~al.(2016)Mostafazadeh, Chambers, He, Parikh, Batra,
  Vanderwende, Kohli, and Allen}]{mostafazadeh2016corpus}
Nasrin Mostafazadeh, Nathanael Chambers, Xiaodong He, Devi Parikh, Dhruv Batra,
  Lucy Vanderwende, Pushmeet Kohli, and James Allen. 2016.
\newblock A corpus and cloze evaluation for deeper understanding of commonsense
  stories.
\newblock In \emph{Proceedings of the 2016 Conference of the North American
  Chapter of the Association for Computational Linguistics: Human Language
  Technologies}, pages 839--849.

\bibitem[{Mostafazadeh et~al.(2017)Mostafazadeh, Roth, Louis, Chambers, and
  Allen}]{mostafazadeh2017lsdsem}
Nasrin Mostafazadeh, Michael Roth, Annie Louis, Nathanael Chambers, and James
  Allen. 2017.
\newblock Lsdsem 2017 shared task: The story cloze test.
\newblock In \emph{Proceedings of the 2nd Workshop on Linking Models of
  Lexical, Sentential and Discourse-level Semantics}, pages 46--51.

\bibitem[{Neiger and Tenenboim-Weinblatt(2016)}]{neiger2016understanding}
Motti Neiger and Keren Tenenboim-Weinblatt. 2016.
\newblock Understanding journalism through a nuanced deconstruction of temporal
  layers in news narratives.
\newblock \emph{Journal of Communication}, 66(1):139--160.

\bibitem[{Nielsen(2015)}]{nielsen2015uncertain}
Rasmus~Kleis Nielsen. 2015.
\newblock The uncertain future of local journalism.
\newblock \emph{Pre-publication version of chapter in Rasmus Kleis Nielsen
  (ed.)}.

\bibitem[{Ning et~al.(2018)Ning, Wu, and Roth}]{ning2018multi}
Qiang Ning, Hao Wu, and Dan Roth. 2018.
\newblock A multi-axis annotation scheme for event temporal relations.
\newblock \emph{arXiv preprint arXiv:1804.07828}.

\bibitem[{Papineni et~al.(2002)Papineni, Roukos, Ward, and
  Zhu}]{papineni2002bleu}
Kishore Papineni, Salim Roukos, Todd Ward, and Wei-Jing Zhu. 2002.
\newblock Bleu: a method for automatic evaluation of machine translation.
\newblock In \emph{Proceedings of the 40th annual meeting of the Association
  for Computational Linguistics}, pages 311--318.

\bibitem[{Peng et~al.(2018)Peng, Ghazvininejad, May, and
  Knight}]{peng2018towards}
Nanyun Peng, Marjan Ghazvininejad, Jonathan May, and Kevin Knight. 2018.
\newblock Towards controllable story generation.
\newblock In \emph{Proceedings of the First Workshop on Storytelling}, pages
  43--49.

\bibitem[{P\"{o}ttker(2003)}]{po2003news}
Horst P\"{o}ttker. 2003.
\newblock News and its communicative quality: the inverted pyramid—when and
  why did it appear?
\newblock \emph{Journalism Studies}, 4(4):501--511.

\bibitem[{Pustejovsky et~al.(2003)Pustejovsky, Hanks, Sauri, See, Gaizauskas,
  Setzer, Radev, Sundheim, Day, Ferro et~al.}]{pustejovsky2003timebank}
James Pustejovsky, Patrick Hanks, Roser Sauri, Andrew See, Robert Gaizauskas,
  Andrea Setzer, Dragomir Radev, Beth Sundheim, David Day, Lisa Ferro, et~al.
  2003.
\newblock The timebank corpus.
\newblock In \emph{Corpus linguistics}, volume 2003, page~40. Lancaster, UK.

\bibitem[{Quan et~al.(2019)Quan, Wang, Le, Yao, Li, and
  Yin}]{quan2019efficient}
Zhe Quan, Zhi-Jie Wang, Yuquan Le, Bin Yao, Kenli Li, and Jian Yin. 2019.
\newblock An efficient framework for sentence similarity modeling.
\newblock \emph{IEEE/ACM Transactions on Audio, Speech, and Language
  Processing}, 27(4):853--865.

\bibitem[{Reimers and Gurevych(2019)}]{reimers2019sentence}
Nils Reimers and Iryna Gurevych. 2019.
\newblock Sentence-bert: Sentence embeddings using siamese bert-networks.
\newblock \emph{arXiv preprint arXiv:1908.10084}.

\bibitem[{Savoy(2013)}]{savoy2013authorship}
Jacques Savoy. 2013.
\newblock Authorship attribution based on a probabilistic topic model.
\newblock \emph{Information Processing \& Management}, 49(1):341--354.

\bibitem[{Scanlan(2003)}]{scanlan2003writing}
Chip Scanlan. 2003.
\newblock Writing from the top down: Pros and cons of the inverted pyramid.
\newblock \emph{Poynter Online., Eri{\c{s}}im tarihi}, 14.

\bibitem[{Shah et~al.(2020)Shah, Schuster, and Barzilay}]{shah2020automatic}
Darsh Shah, Tal Schuster, and Regina Barzilay. 2020.
\newblock Automatic fact-guided sentence modification.
\newblock In \emph{Proceedings of the AAAI Conference on Artificial
  Intelligence}, volume~34, pages 8791--8798.

\bibitem[{Shahaf and Guestrin(2010)}]{shahaf2010connecting}
Dafna Shahaf and Carlos Guestrin. 2010.
\newblock Connecting the dots between news articles.
\newblock In \emph{Proceedings of the 16th ACM SIGKDD international conference
  on Knowledge discovery and data mining}, pages 623--632.

\bibitem[{Shen et~al.(2019)Shen, Zhang, Zhao, Liang, Jin, and
  Wang}]{shen2019intellimerge}
Bo~Shen, Wei Zhang, Haiyan Zhao, Guangtai Liang, Zhi Jin, and Qianxiang Wang.
  2019.
\newblock Intellimerge: a refactoring-aware software merging technique.
\newblock \emph{Proceedings of the ACM on Programming Languages},
  3(OOPSLA):1--28.

\bibitem[{Shen et~al.(2017)Shen, Lin, Tu, Zhao, Liu, Sun
  et~al.}]{shen2017recent}
Shi-Qi Shen, Yan-Kai Lin, Cun-Chao Tu, Yu~Zhao, Zhi-Yuan Liu, Mao-Song Sun,
  et~al. 2017.
\newblock Recent advances on neural headline generation.
\newblock \emph{Journal of computer science and technology}, 32(4):768--784.

\bibitem[{Silva and Valente(2017)}]{silva2017refdiff}
Danilo Silva and Marco~Tulio Valente. 2017.
\newblock Refdiff: detecting refactorings in version histories.
\newblock In \emph{2017 IEEE/ACM 14th International Conference on Mining
  Software Repositories (MSR)}, pages 269--279. IEEE.

\bibitem[{Spangher et~al.(2020)Spangher, May, Ferrara, and
  Peng}]{spangher2020sourcefinding}
Alexander Spangher, Jonathan May, Emilio Ferrara, and Nanyun Peng. 2020.
\newblock ``don't quote me on that'': Finding mixtures of sources in news
  articles.
\newblock In \emph{Proceedings of Computation+Journalism Conference}.

\bibitem[{Spangher et~al.(2021{\natexlab{a}})Spangher, May, Shiang, and
  Deng}]{spangher2021multitask}
Alexander Spangher, Jonathan May, Sz-Rung Shiang, and Lingjia Deng.
  2021{\natexlab{a}}.
\newblock \href {https://doi.org/10.18653/v1/2021.emnlp-main.40} {Multitask
  semi-supervised learning for class-imbalanced discourse classification}.
\newblock In \emph{Proceedings of the 2021 Conference on Empirical Methods in
  Natural Language Processing}, pages 498--517, Online and Punta Cana,
  Dominican Republic. Association for Computational Linguistics.

\bibitem[{Spangher et~al.(2021{\natexlab{b}})Spangher, Scott, and
  Huang-Isherwood}]{spangher2021annenberg}
Alexander Spangher, Amberg-Lynn Scott, and Ke~Huang-Isherwood.
  2021{\natexlab{b}}.
\newblock \href {https://www.instagram.com/p/CNsUlUflYV0/} {``what's the
  diff?'': Examining news article updates and changing narratives during the
  uss theodore roosevelt coronavirus crisis}.
\newblock In \emph{Annenberg Scymposium}.

\bibitem[{Stahl(2006)}]{stahl2006difference}
Bernd~Carsten Stahl. 2006.
\newblock On the difference or equality of information, misinformation, and
  disinformation: A critical research perspective.
\newblock \emph{Informing Science}, 9.

\bibitem[{Tamori et~al.(2017)Tamori, Hitomi, Okazaki, and
  Inui}]{tamori2017analyzing}
Hideaki Tamori, Yuta Hitomi, Naoaki Okazaki, and Kentaro Inui. 2017.
\newblock \href {https://doi.org/10.18653/v1/W17-4208} {Analyzing the revision
  logs of a {J}apanese newspaper for article quality assessment}.
\newblock In \emph{Proceedings of the 2017 {EMNLP} Workshop: Natural Language
  Processing meets Journalism}, pages 46--50, Copenhagen, Denmark. Association
  for Computational Linguistics.

\bibitem[{Tan and Bockisch(2019)}]{tan2019survey}
Liang Tan and Christoph Bockisch. 2019.
\newblock A survey of refactoring detection tools.
\newblock In \emph{Software Engineering (Workshops)}, pages 100--105.

\bibitem[{Tian et~al.(2020{\natexlab{a}})Tian, Chakrabarty, Morstatter, and
  Peng}]{tian2020identifying}
Yufei Tian, Tuhin Chakrabarty, Fred Morstatter, and Nanyun Peng.
  2020{\natexlab{a}}.
\newblock Identifying cultural differences through multi-lingual wikipedia.
\newblock \emph{arXiv preprint arXiv:2004.04938}.

\bibitem[{Tian et~al.(2020{\natexlab{b}})Tian, Zhang, Liu, Zhao, Jia, and
  Sheng}]{tian2020scene}
Zhixing Tian, Yuanzhe Zhang, Kang Liu, Jun Zhao, Yantao Jia, and Zhicheng
  Sheng. 2020{\natexlab{b}}.
\newblock Scene restoring for narrative machine reading comprehension.
\newblock In \emph{Proceedings of the 2020 Conference on Empirical Methods in
  Natural Language Processing (EMNLP)}, pages 3063--3073.

\bibitem[{Tsantalis et~al.(2018)Tsantalis, Mansouri, Eshkevari, Mazinanian, and
  Dig}]{tsantalis2018accurate}
Nikolaos Tsantalis, Matin Mansouri, Laleh Eshkevari, Davood Mazinanian, and
  Danny Dig. 2018.
\newblock Accurate and efficient refactoring detection in commit history.
\newblock In \emph{2018 IEEE/ACM 40th International Conference on Software
  Engineering (ICSE)}, pages 483--494. IEEE.

\bibitem[{Usher(2018)}]{usher2018breaking}
Nikki Usher. 2018.
\newblock Breaking news production processes in us metropolitan newspapers:
  Immediacy and journalistic authority.
\newblock \emph{Journalism}, 19(1):21--36.

\bibitem[{Van~Dijk(1983)}]{van1983discourse}
Teun~A Van~Dijk. 1983.
\newblock Discourse analysis: Its development and application to the structure
  of news.
\newblock \emph{Journal of communication}, 33(2):20--43.

\bibitem[{Walker et~al.(2006)Walker, Strassel, Medero, and Maeda}]{ace2005}
Christopher Walker, Stephanie Strassel, Julie Medero, and Kazuaki Maeda. 2006.
\newblock {ACE} 2005 multilingual training corpus {LDC2006T06}.
\newblock Linguistic Data Consortium, Philadelphia.

\bibitem[{Wang et~al.(2020)Wang, Matsumura, Correnti, Litman, Zhang, Howe,
  Magooda, and Quintana}]{wang2020erevis}
Elaine~Lin Wang, Lindsay~Clare Matsumura, Richard Correnti, Diane Litman,
  Haoran Zhang, Emily Howe, Ahmed Magooda, and Rafael Quintana. 2020.
\newblock {eRevis(ing)}: Students’ revision of text evidence use in an
  automated writing evaluation system.
\newblock \emph{Assessing Writing}, 44:100449.

\bibitem[{Wolf et~al.(2020)Wolf, Debut, Sanh, Chaumond, Delangue, Moi, Cistac,
  Rault, Louf, Funtowicz, Davison, Shleifer, von Platen, Ma, Jernite, Plu, Xu,
  Le~Scao, Gugger, Drame, Lhoest, and Rush}]{wolf2019huggingface}
Thomas Wolf, Lysandre Debut, Victor Sanh, Julien Chaumond, Clement Delangue,
  Anthony Moi, Pierric Cistac, Tim Rault, Remi Louf, Morgan Funtowicz, Joe
  Davison, Sam Shleifer, Patrick von Platen, Clara Ma, Yacine Jernite, Julien
  Plu, Canwen Xu, Teven Le~Scao, Sylvain Gugger, Mariama Drame, Quentin Lhoest,
  and Alexander Rush. 2020.
\newblock \href {https://doi.org/10.18653/v1/2020.emnlp-demos.6} { 
Transformers: State-of-the-art natural language processing}.
\newblock In \emph{Proceedings of the 2020 Conference on Empirical Methods in
  Natural Language Processing: System Demonstrations}, pages 38--45, Online.
  Association for Computational Linguistics.

\bibitem[{Wu et~al.(2020)Wu, Kanoulas, de~Rijke, and Lu}]{wu2020wn}
Chuan Wu, Evangelos Kanoulas, Maarten de~Rijke, and Wei Lu. 2020.
\newblock Wn-salience: A corpus of news articles with entity salience
  annotations.
\newblock In \emph{Proceedings of The 12th Language Resources and Evaluation
  Conference}, pages 2095--2102.

\bibitem[{Wu et~al.(2022)Wu, Spangher, Alipoormolabashi, Freedman, Weischedel,
  and Peng}]{wu2022procedural}
Te-Lin Wu, Alex Spangher, Pegah Alipoormolabashi, Marjorie Freedman, Ralph
  Weischedel, and Nanyun Peng. 2022.
\newblock Understanding multimodal procedural knowledge by sequencing
  multimodal instructional manuals.
\newblock In \emph{Proceedings of the Conference of the 60th Annual Meeting of
  the Association for Computational Linguistics (ACL)}.

\bibitem[{Xu et~al.(2019)Xu, Yao, Xu, Gu, Tong, and Lu}]{xu2019commit}
Shengbin Xu, Yuan Yao, Feng Xu, Tianxiao Gu, Hanghang Tong, and Jian Lu. 2019.
\newblock Commit message generation for source code changes.
\newblock In \emph{IJCAI}.

\bibitem[{Yang et~al.(2017)Yang, Halfaker, Kraut, and
  Hovy}]{yang2017identifying}
Diyi Yang, Aaron Halfaker, Robert Kraut, and Eduard Hovy. 2017.
\newblock Identifying semantic edit intentions from revisions in wikipedia.
\newblock In \emph{Proceedings of the 2017 Conference on Empirical Methods in
  Natural Language Processing}, pages 2000--2010.

\bibitem[{Yao et~al.(2018)Yao, Liu, and Zhang}]{yao2018novel}
Haipeng Yao, Huiwen Liu, and Peiying Zhang. 2018.
\newblock A novel sentence similarity model with word embedding based on
  convolutional neural network.
\newblock \emph{Concurrency and Computation: Practice and Experience},
  30(23):e4415.

\bibitem[{Yao et~al.(2019)Yao, Peng, Weischedel, Knight, Zhao, and
  Yan}]{yao2019plan}
Lili Yao, Nanyun Peng, Ralph Weischedel, Kevin Knight, Dongyan Zhao, and Rui
  Yan. 2019.
\newblock Plan-and-write: Towards better automatic storytelling.
\newblock In \emph{Proceedings of the AAAI Conference on Artificial
  Intelligence}, volume~33, pages 7378--7385.

\bibitem[{Yatskar et~al.(2010)Yatskar, Pang, Danescu-Niculescu-Mizil, and
  Lee}]{yatskar2010sake}
Mark Yatskar, Bo~Pang, Cristian Danescu-Niculescu-Mizil, and Lillian Lee. 2010.
\newblock \href {https://aclanthology.org/N10-1056} {For the sake of
  simplicity: Unsupervised extraction of lexical simplifications from
  {W}ikipedia}.
\newblock In \emph{Human Language Technologies: The 2010 Annual Conference of
  the North {A}merican Chapter of the Association for Computational
  Linguistics}, pages 365--368, Los Angeles, California. Association for
  Computational Linguistics.

\bibitem[{Yimam et~al.(2017)Yimam, {\v{S}}tajner, Riedl, and
  Biemann}]{yimam2017cwig3g2}
Seid~Muhie Yimam, Sanja {\v{S}}tajner, Martin Riedl, and Chris Biemann. 2017.
\newblock Cwig3g2-complex word identification task across three text genres and
  two user groups.
\newblock In \emph{Proceedings of the Eighth International Joint Conference on
  Natural Language Processing (Volume 2: Short Papers)}, pages 401--407.

\bibitem[{Yin et~al.(2018)Yin, Neubig, Allamanis, Brockschmidt, and
  Gaunt}]{yin2018learning}
Pengcheng Yin, Graham Neubig, Miltiadis Allamanis, Marc Brockschmidt, and
  Alexander~L Gaunt. 2018.
\newblock Learning to represent edits.
\newblock \emph{arXiv preprint arXiv:1810.13337}.

\bibitem[{Zanzotto and Pennacchiotti(2010)}]{zanzotto2010expanding}
Fabio~Massimo Zanzotto and Marco Pennacchiotti. 2010.
\newblock Expanding textual entailment corpora from wikipedia using
  co-training.
\newblock In \emph{Proceedings of the 2nd Workshop on The People’s Web Meets
  NLP: Collaboratively Constructed Semantic Resources}, pages 28--36.

\bibitem[{Zhang et~al.(2017)Zhang, Hashemi, Hwa, and Litman}]{zhang2017corpus}
Fan Zhang, Homa~B Hashemi, Rebecca Hwa, and Diane Litman. 2017.
\newblock A corpus of annotated revisions for studying argumentative writing.
\newblock In \emph{Proceedings of the 55th Annual Meeting of the Association
  for Computational Linguistics (Volume 1: Long Papers)}, pages 1568--1578.

\bibitem[{Zhang and Litman(2015)}]{zhang2015annotation}
Fan Zhang and Diane Litman. 2015.
\newblock \href {https://doi.org/10.3115/v1/W15-0616} {Annotation and
  classification of argumentative writing revisions}.
\newblock In \emph{Proceedings of the Tenth Workshop on Innovative Use of {NLP}
  for Building Educational Applications}, pages 133--143, Denver, Colorado.
  Association for Computational Linguistics.

\bibitem[{Zhang and Wan(2017)}]{zhang2017towards}
Jianmin Zhang and Xiaojun Wan. 2017.
\newblock Towards automatic construction of news overview articles by news
  synthesis.
\newblock In \emph{Proceedings of the 2017 Conference on Empirical Methods in
  Natural Language Processing}, pages 2111--2116.

\bibitem[{Zhang(2020)}]{zhang2020engaging}
Zhe~Victor Zhang. 2020.
\newblock Engaging with automated writing evaluation ({AWE}) feedback on {L2}
  writing: Student perceptions and revisions.
\newblock \emph{Assessing Writing}, 43:100439.

\end{thebibliography}
\clearpage
\newpage

\newpage

\appendix

\section{Dataset: Broader Scope}
\label{app:broader_scope}

We expect that \textit{NewsEdits}
 will be useful for a range of existing tasks for revision corpora, such as edit language modeling \cite{yin2018learning} and grammatical error correction \cite{grundkiewicz2014wiked}. We also think \textit{NewsEdits} can impact other areas of NLP research and computational journalism, including:

\begin{enumerate}
\item \textbf{Resource Allocation in Newsrooms} Newsrooms are often tasked with covering multiple breaking news stories that are unfolding simultanesouly \cite{usher2018breaking}. When multiple stories are being published to cover breaking news, or multiple news events are breaking at the same time, newsrooms are often forced to make decisions on which journalists to assign to continue reporting stories. This becomes especially pronounced in an era of budget cuts and local-journalism shortages \cite{nielsen2015uncertain}. We interviewed 3 journalists with over 20 years of experience at major breaking news outlets. They agreed that a predictive system that performed the tasks explored in Section \ref{sct:tasks} would be very helpful for allowing editors track which stories are most likely to change the most, allowing them to keep resources on these stories.

\item\textbf{Event-temporal relation extraction} \cite{ning2018multi} and \textbf{Fact-guided updates} \cite{shah2020automatic}. As shown in Tables \ref{tab:add_sentence_attr} and \ref{tab:event_updates}, added and edited sentences are both more likely to contain events, and event updates. We see potential for using these sentences to train revise-and-edit \cite{hashimoto2018retrieve} models.
   
\item \textbf{Misinformation}: Journalists often issue formal \textit{Corrections} when they discover errors in their reporting \cite{appelman2015news}\footnote{An example of \textit{misinformation} vs. \textit{disinformation} \cite{stahl2006difference}}. We found 14,301 corrections in \textit{added} sentences across the same sample with a custom lexicon\footnote{In other words, the corrections were \textit{not} present in previous drafts of the article. See Appendix \ref{app:corrections} for examples.}. This might be used to help compare malicious campaigns with honest errors \cite{ferrara2017disinformation}.
 
\item \textbf{Headline Generation} \cite{shen2017recent}. Across a sample of 2 million version pairs, we count 376,944, or 17\% that have a headline update. Headlines have been used to predict emotional salience \cite{gupta2019predicting}. Modeling edits that result in headline changes can help differentiate salient from non-salient edits. 

\item \textbf{Authorship Attribution} is the task of predicting which authors were involved in writing an article. We found 2,747 \textit{Contributor Lines}\footnote{Contribution acknowledgement. Appendix \ref{app:corrections} for ex.} added to articles. This can provide a temporal extension to author-attribution models such as \newcite{savoy2013authorship}.

\item \textbf{Identifying Informational Needs}: Source inclusion \cite{spangher2020sourcefinding} and discourse structures \cite{choubey2020discourse, spangher2021multitask} of static articles have been studied. We see this corpus as being useful for studying \textit{when} these narrative elements are added. 
\end{enumerate}

Directions that we have not explored, but possibly interesting include: style transfer \cite{fu2018style}, detecting bias in news articles \cite{mehrabi2020man}, cross-cultural sensitivity \cite{tian2020identifying}, insertion-based article generation \cite{lu2021efficient}, and framing changes in response to an unfolding story \cite{spangher2021annenberg}.




\section{Exploratory Analysis Details}
\label{app:eda_details}

Insight \#2 in Section \ref{sct:eda} was based on several experiments that we ran. Here we provide more details about the experiments we ran.

\noindent\textbf{Events:} We sample of 200,000 documents (7 million sentences) from our corpus\footnote{We balance for newspaper source, article length (from 5 to 100 sentences), and number of additions/deletions (from 0\% of article to 50\%)} and use Eventplus \cite{ma2021eventplus} to extract all events. We find added/deleted sentences have significantly more events than unchanged sentences.

\noindent\textbf{Quotes:} Using a quote extraction pipeline \cite{spangher2020sourcefinding}, we extract explicit and implicit quotes from the sample of documents used above. The pipeline identifies patterns associated with quotes (e.g. double quotation marks) to distantly supervise training an algorithm to extract a wide variety of implicit and explicit quotes with high accuracy ($.8$ F1-score). We find added/deleted sentences contain significantly more quotes than unchanged sentences.

\noindent\textbf{News Discourse:} We train a model to identify three coarse-grained discourse categories in news text: \textit{Main} (i.e. main story)  \textit{Cause} (i.e. immediate context), and \textit{Distant} (i.e. history, analysis, etc.) We use a news discourse schema \cite{van1983discourse} and a labeled dataset which contains 800 news articles labeled on the sentence-level \cite{choubey2020discourse}. We train a model on this dataset to score news articles in our dataset\footnote{We achieve a macro F1-score of .67 on validation data using the architecture described in \newcite{spangher2021multitask}.}. \textit{Then}, we filter to \textit{Addition}, \textit{Deletion}, etc. sentences. We show that added and deleted sentences are significantly more likely than unchanged sentences to be \textit{Main} or \textit{Cause} sentences, while unchanged sentences are significantly more likely to be \textit{Distant}.

\section{Error Analysis: Continued}
\label{app:error_analysis}
\begin{table*}
\centering
\small
\begin{tabular}{p{1.3cm}p{1.5cm}p{1.3cm}p{1.3cm}p{1.3cm}p{1.3cm}p{1.3cm}p{1.3cm}p{1.3cm}}
\toprule
U.S. \quad Politics & U.K.\quad\quad Politics &  Police \textit{Crime}  & Aviation  & Tragedy & War & Criminals \textit{Crime} & School & Violence \textit{Crime} \\
(topic 0) &  (topic 2) &  (topic 5)  & (topic 6)  &  (topic 7) & (topic 9) & (topic 12) & (topic 13) & (topic 18) \\
\midrule
       mr & government &    police &    people &   family &   killed &    court &   school &   police \\
president &      party &       man &   airport &     died &   people &     year &     year & officers \\
    trump &         mr &       old &     plane & hospital &   attack &      old &    world &   people \\
 minister &     labour &      year &  aircraft &      old &       al &       mr &      new &     area \\
    prime &    council &  arrested &  reported &      man &   forces &      man &   people & incident \\
    house &   minister &     woman &    agency &  service &  attacks &   murder &     city &    local \\
   donald &     leader &  officers & officials &   rescue &    group &   police &     time &    scene \\
    obama &        new &       men &      news &     year & military &    years &    years &     shot \\
    white &     people & suspicion &       air &   police &     city &     told &      day & shooting \\
      new &  secretary &    london &    flight &    death & security &   guilty &    event &  injured \\
\bottomrule
\end{tabular}
\caption{Topic Model: Top Topics, selected on the bases of the number of documents they are most-expressed in. Labels are assigned by the researchers post-hoc. Several topics appear to be subsets of a broader \textit{Crime} topic: we note the superclass \textit{Crime} in parentheses. The specific \textit{Crime} topic mentioned in the main body is the Violence topic (Topic 18)}
\label{tbl:topic_model}
\end{table*}

As discussed in Section \ref{sct:error_analysis}, we perform Latent Dirichlet Allocation \cite{blei2003latent} to soft-cluster documents. In Table \ref{tbl:topic_model}, we show the top $k=10$ words for each topic $i$ (i.e. $\beta_{1,...k}^{i}$ where $\beta_{1}^{i} > \beta_{2}^{i} > ... >  \beta_{k}^{i}$). 

\section{Experiment Details}
\label{app:exp_details}

\subsection{Modeling Decisions}

For \textbf{Task 1}, we sample documents in our training dataset, balancing across versions and $y$ and exclude articles with more than 6,000 characters. However, because of the imbalanced nature of the dataset, we could not fully balance. 

\begin{figure}
    \centering
    \includegraphics[width=.6\linewidth]{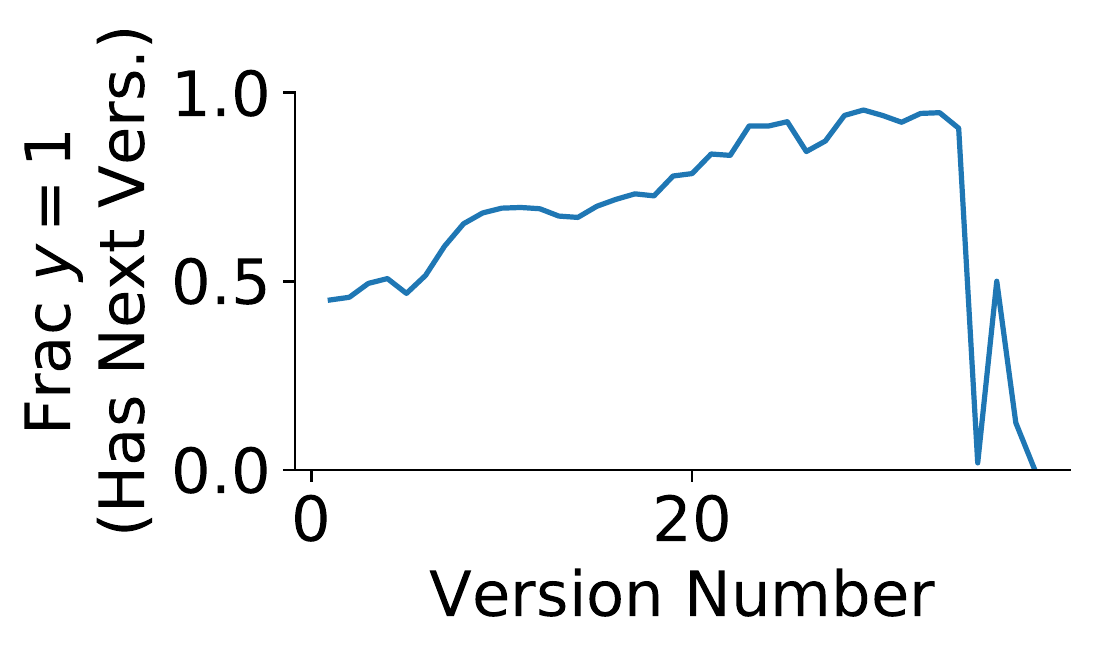}
    \caption{Percentage of the training dataset for \textbf{Task 1} which contains $y=1$, or where another version of the article has been published.}
    \label{fig:task_1_details}
\end{figure}

As is seen in Table \ref{fig:versions_per_article}, +Version, the version number of the old version had a large effect on the performance of the model, boosting performance by over 10 points. We believe that this is permissible, because the version number of the old article is available at prediction time. Interestingly, the effect is actually the opposite of what we would expect. As can be seen in Figure \ref{fig:task_1_details}, the more versions an article has, the \textit{more likely} it is to contain another version. This is perhaps because articles with many versions are \textit{breaking news} articles, and they behave differently than articles with fewer versions. To more properly test a model's ability to judge breaking news specifically, we can create a validation set where all versions of a set of articles are included; thus the model is forced to identify at early versions whether an article is a breaking news story or not.

For \textbf{Task 2}, we first experiment with different regression modeling heads before reframing the task as a classification task. We test with Linear Regression and Poisson Regression, seeking to learn the raw counts. However, we found that we were not able to improve above random in any subtask and reframed the problem as a binned classification problem.

\subsection{Hyperparameters and Training}
\label{app:exp_details:hyperparameters}

For all tasks, we used pretrained \texttt{RoBERTa Base} from \newcite{wolf2019huggingface}.  We used reasonable defaults for learning rate, dropout and other hyperparameters explored in \newcite{spangher2021multitask}, which we describe now. For all tasks, we used AdamW as an optimizer, with values $\beta_1 = .9$, $\beta_2 = .99$, $\epsilon = 1e-8$. We used batch-size $=1$ but experimented with different gradient accumulations (i.e. effective batch size) $\in [10, 20, 100]$. We did not find much impact to varying this parameter. We used a learning rate of 1e-6 as in \newcite{spangher2021multitask}. Early in experimentation, we trained for 10 epochs, but did not observe any improvement past the 3rd epoch, so we limited training to 5 epochs. We used a dropout probability of $.1$, $0$ warmup steps and $0$ weight decay. The embedding dimensionality for the pretrained \texttt{RoBERTa Base} we used is 768, and for all other layers, we used a hidden-dimension of 512.

For deriving sentence embeddings, we tested several different methods. We tested both using the \texttt{<sep>} token from RoBERTa and averaging the word-embeddings of each word-piece, as in \newcite{spangher2021multitask}, but found that a third method---using self-attention over the word embeddings, or a learned, weighted average---performed the best. We concatenated a sentence-level positional embedding vector, as in \newcite{spangher2021multitask}, with a max cutoff of $40$ positional embeddings (i.e. every sentence with an index greater than 40 was assigned the same vector.)

\section{Dataset Details}
\label{app:dataset_details}

Here, we give additional details on the dataset, starting with relevant analyses and ending with technical details that should guide the user on how to access our dataset. 

\subsection{Additional Analysis}

\begin{figure}[t]
\centering
\subfloat[Distribution over days per update, group 1. Median across all sources in this group is 21 days.]{%
  \includegraphics[width=.48\linewidth]{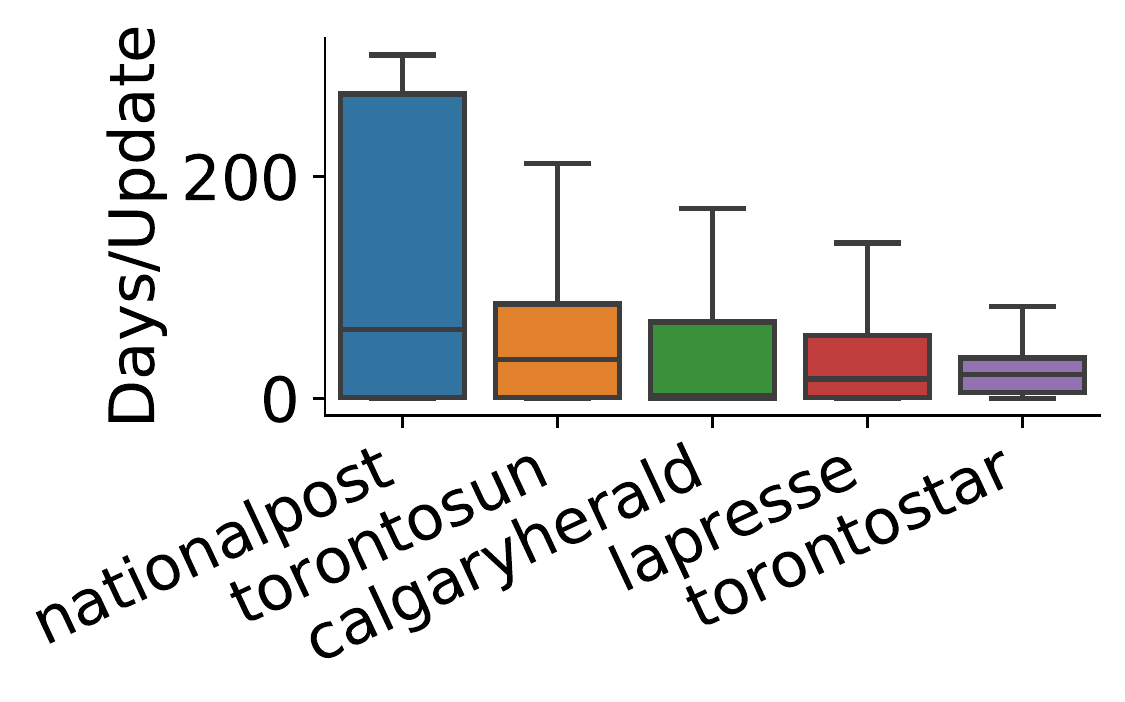}%
  \label{fig:update_time:1}%
}\hspace{.04cm}
\subfloat[Distribution over days per update, group 2. Median across all sources in this group is .9 days]{%
  \includegraphics[width=.48\linewidth]{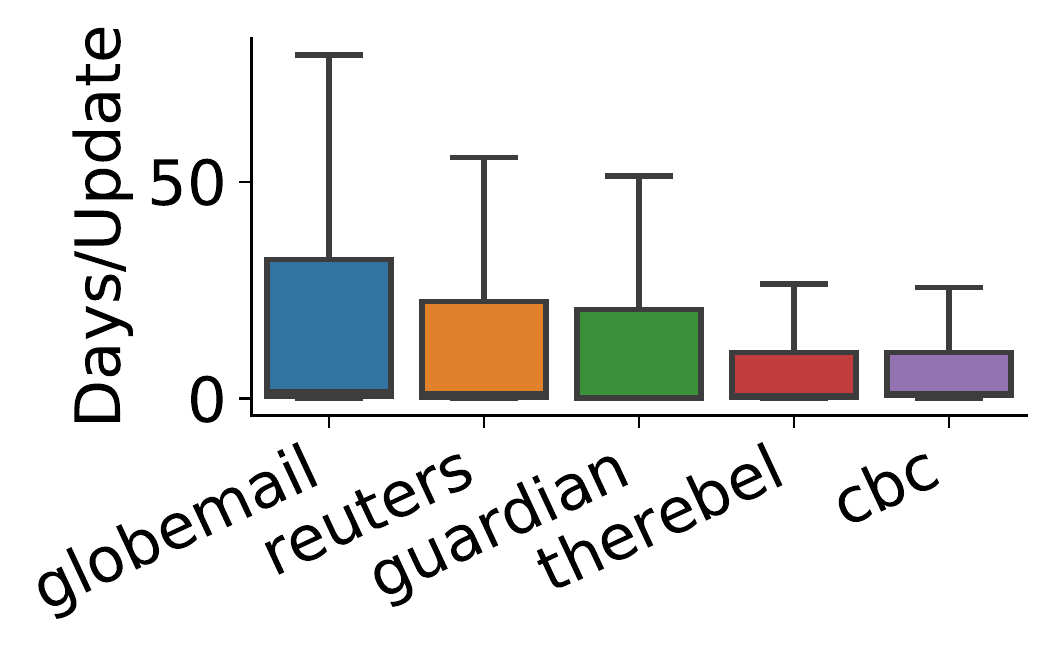}%
  \label{fig:update_time:2}%
}\hspace{.05cm}
\subfloat[Distribution over days per update, group 3. Median across all sources in this group is .35 days, or 8.4 hours]{%
  \includegraphics[width=.48\linewidth]{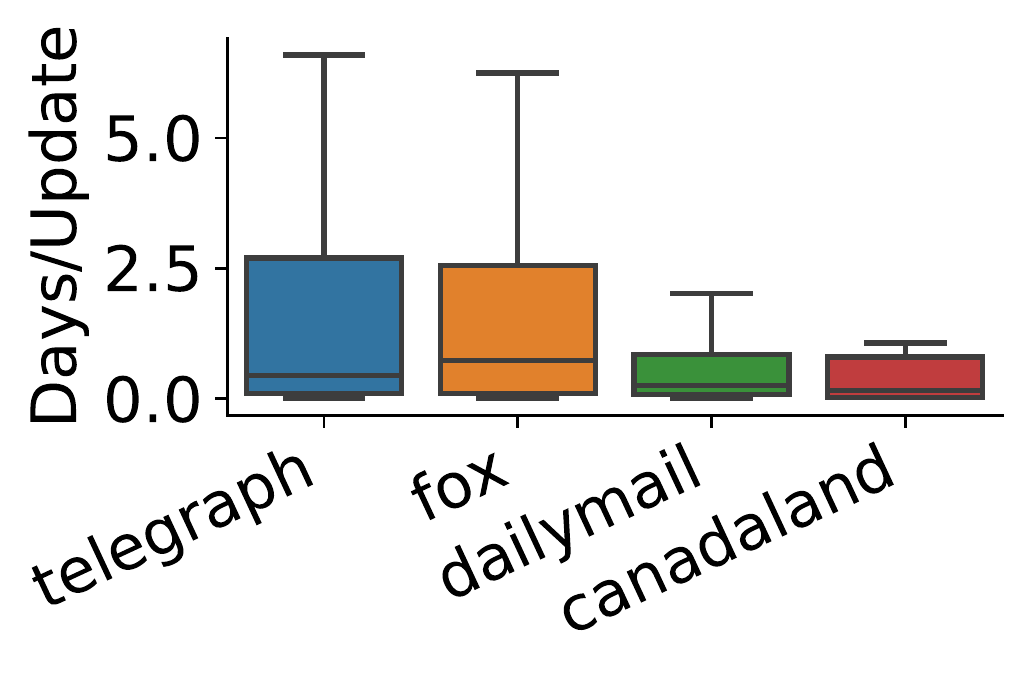}%
  \label{fig:update_time:3}%
}\hspace{.04cm}
\subfloat[Distribution over days per update, group 4. Median across all sources in this group is .05 days, or 1.33 hours.]{%
  \includegraphics[width=.48\linewidth]{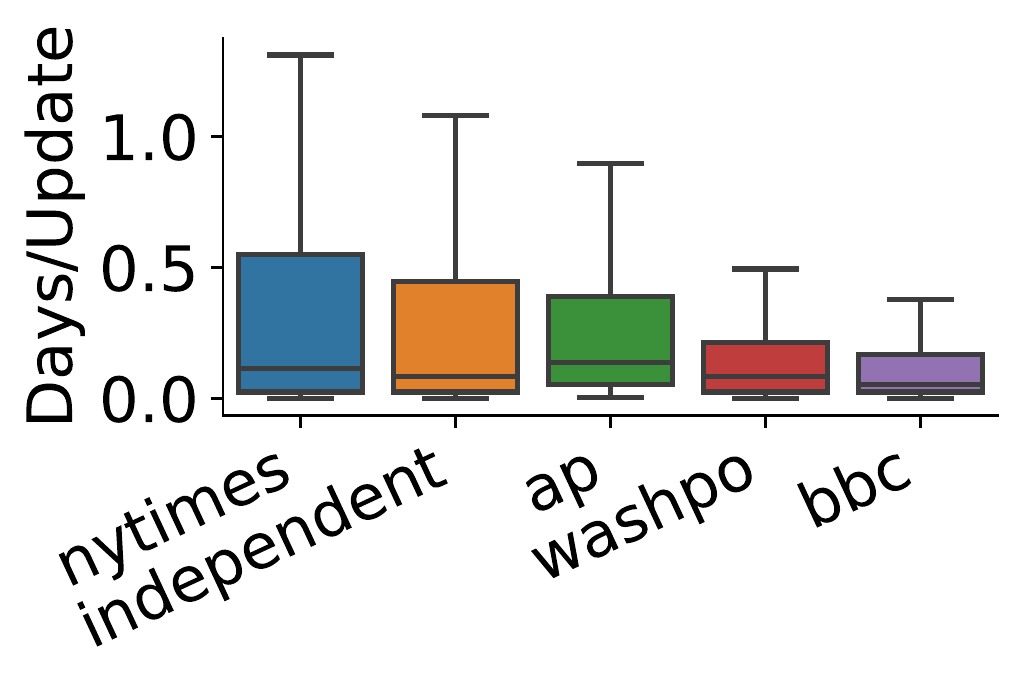}%
  \label{fig:update_time:4}%
}
\caption{Average time between version updates. We break sources into four primary groups with similar update distributions.}
\label{fig:update_time}
\end{figure}

\begin{figure}[t]
\centering
\subfloat[Distributions over discourse tags, by article length.]{%
  \includegraphics[width=.48\linewidth]{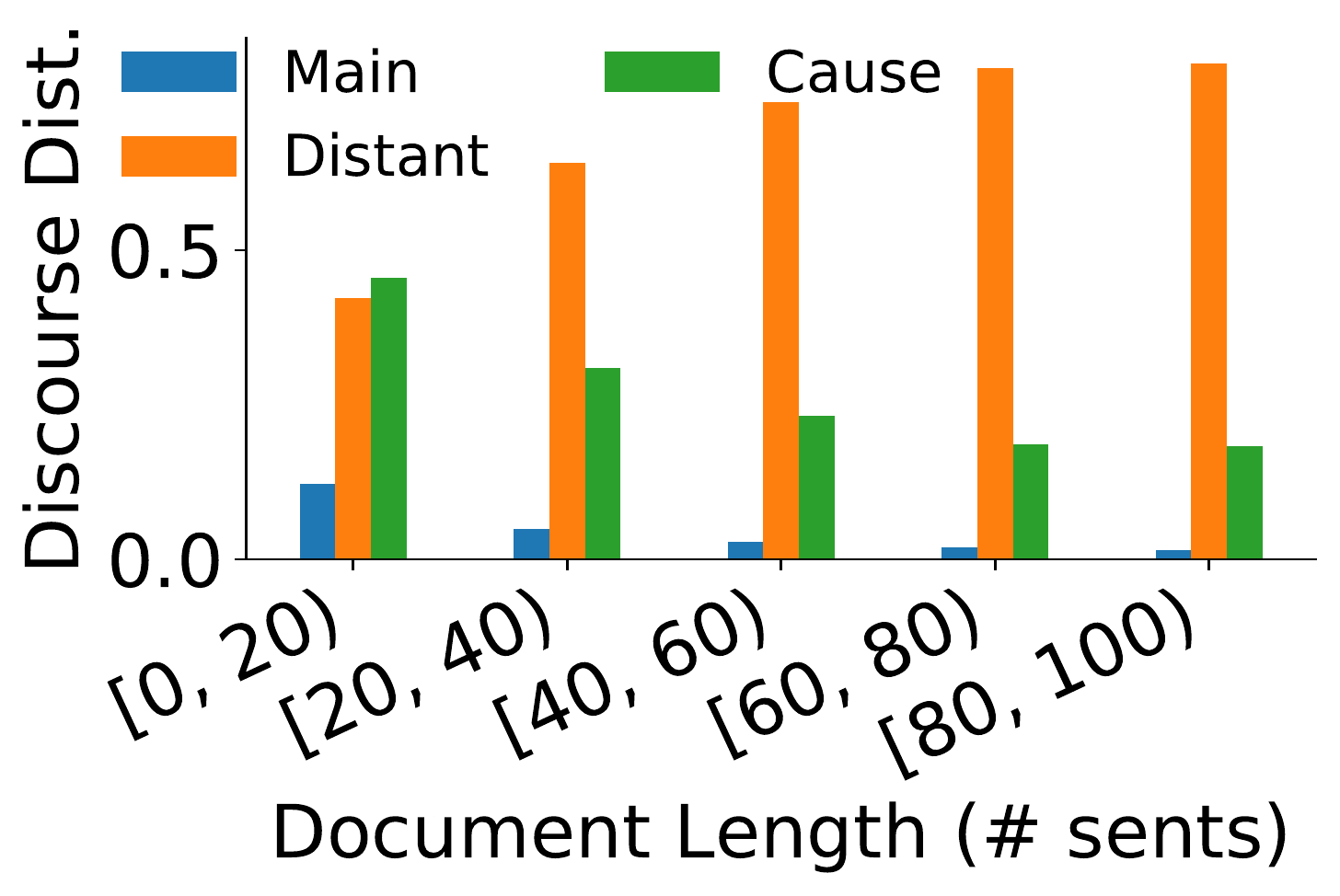}%
  \label{fig:discourse_dynamics:dist_by_len}%
}\hspace{.04cm}
\subfloat[Odds of discourse element, by version. $odds = p(d | v) / p(d| !v)$.]{%
  \includegraphics[width=.48\linewidth]{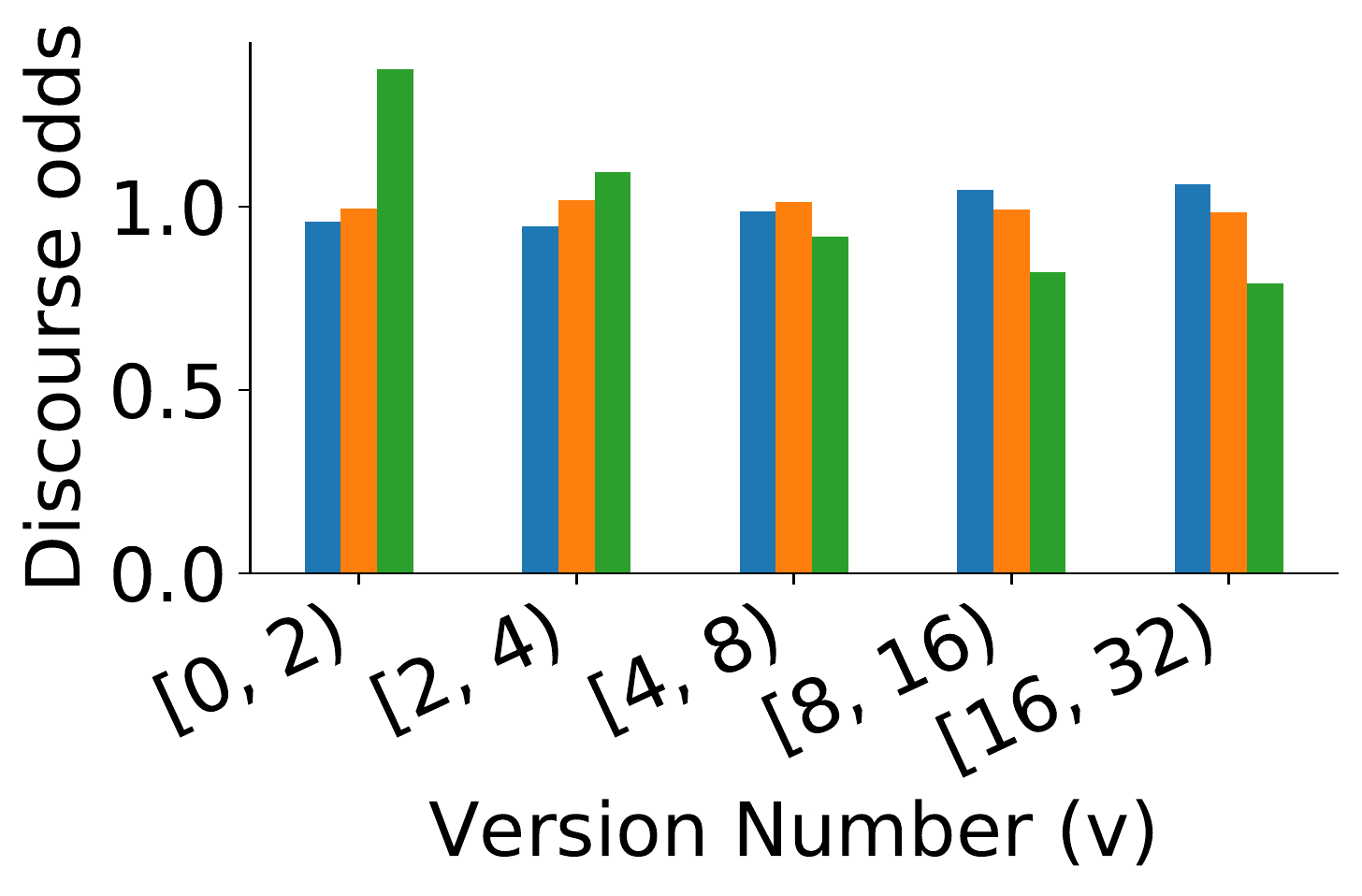}%
  \label{fig:discourse_dynamics:odds_by_version}%
}\hspace{.05cm}
\subfloat[Odds of discourse element in added sentences. $odds = p(d | v, a) / p(d | v, !a)$.]{%
  \includegraphics[width=.48\linewidth]{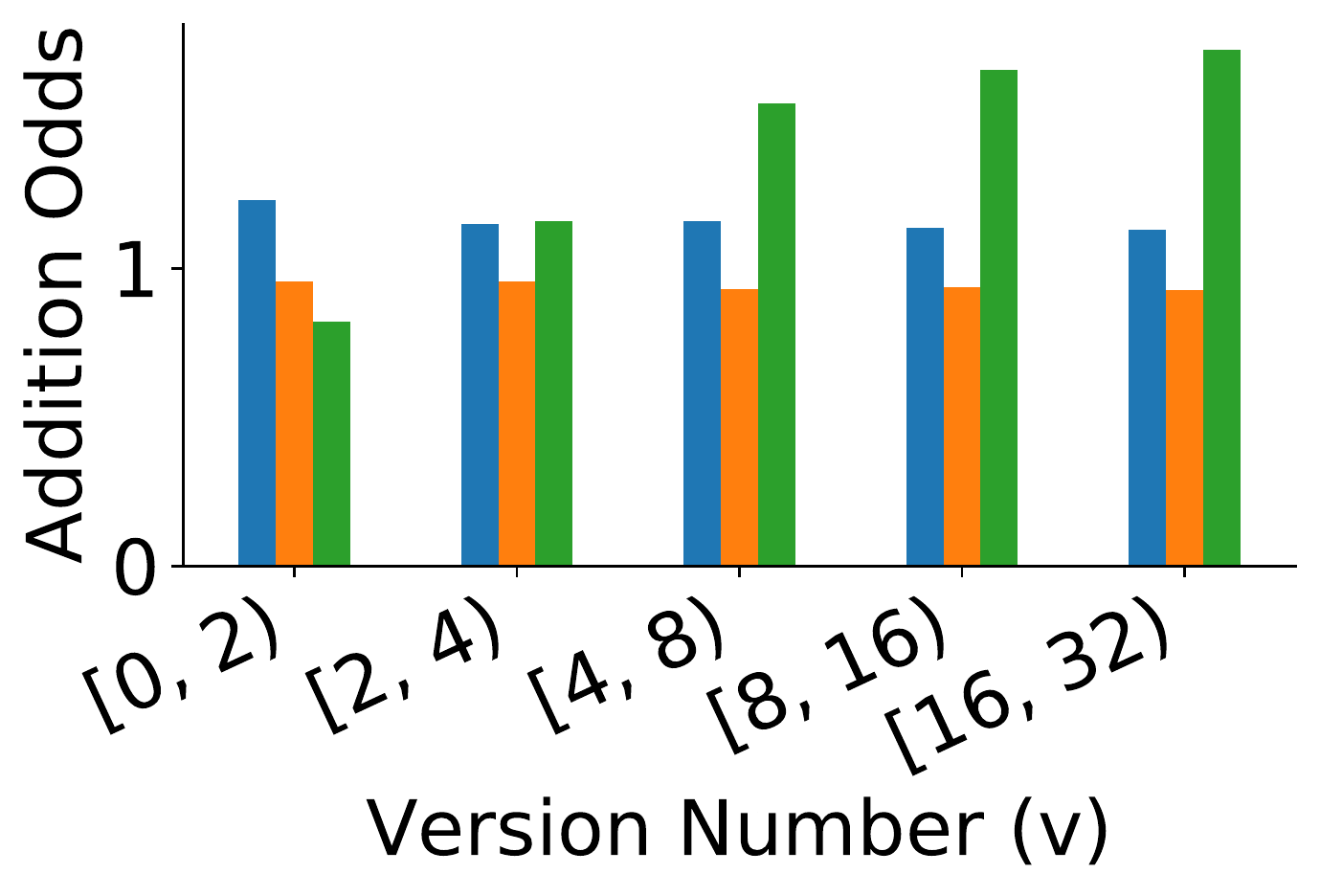}%
  \label{fig:discourse_dynamics:addition_odds}%
}\hspace{.04cm}
\subfloat[Odds of discourse element in deleted sentences. $odds = p(d | v, del) / p(d | v, !del)$.]{%
  \includegraphics[width=.48\linewidth]{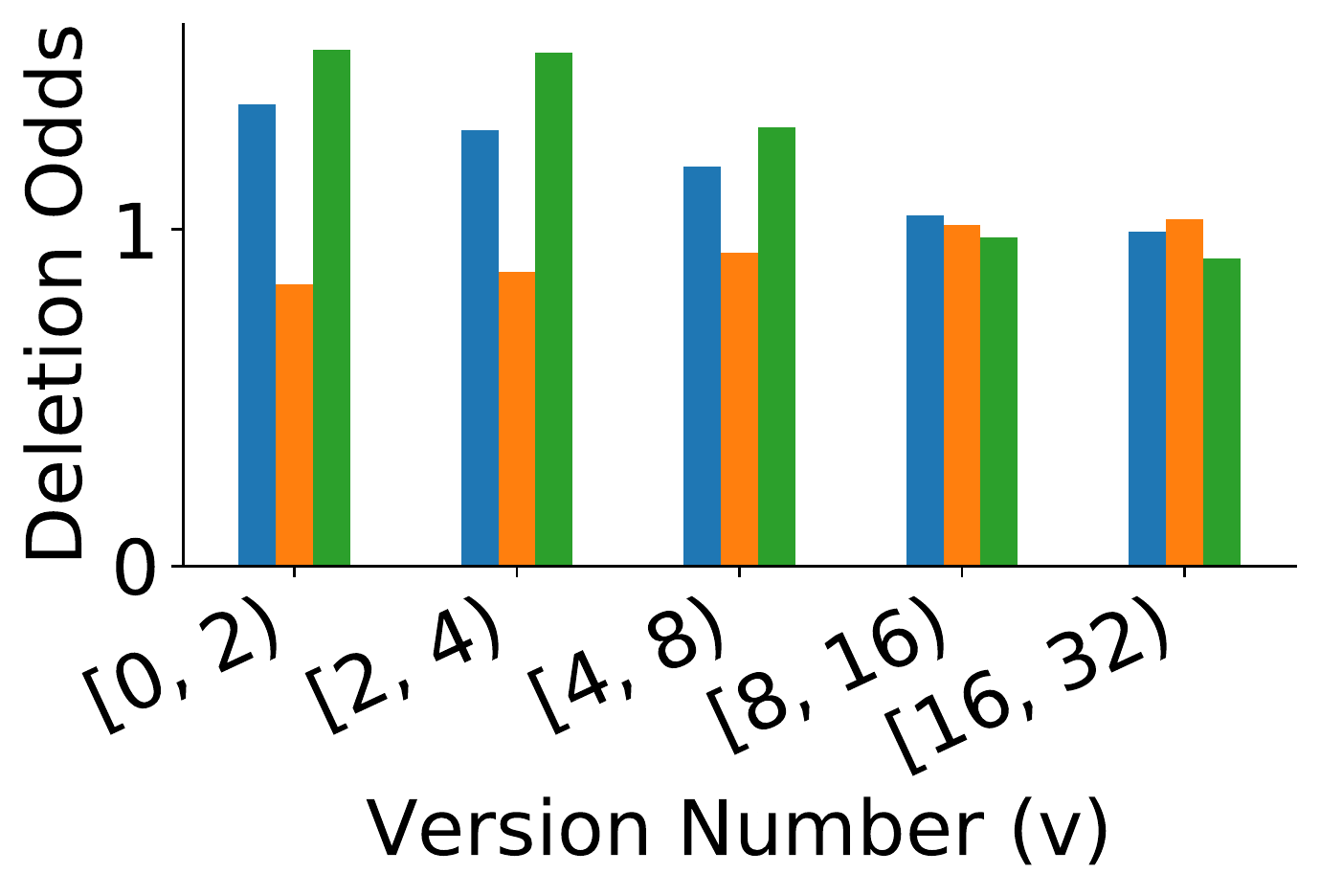}%
  \label{fig:discourse_dynamics:deletion_odds}%
}
\caption{Dynamics of news discourse composition size across time. $d$ refers to \textit{discourse label}, $v$ refers to \textit{version} and $a$, $del$ refer to \textit{is\_added}, \textit{is\_deleted}}
\end{figure}

\subsubsection{Amount of time between Versions}

The amount of time between republication of an article varies widely across news outlets, and has a large role in determining what kinds of stories are being republished. As can be seen in Figure \ref{fig:update_time}, we group sources into 4 categories: (1) Figure \ref{fig:update_time:1}, those that update articles over weeks (tabloids and magazines), (2) Figure \ref{fig:update_time:2}, those that update articles on a daily basis, on median, (3) Figure \ref{fig:update_time:3}, those that update 2-3 times a day, and (4) Figure \ref{fig:update_time:4}, those that update hourly, or breaking news outlets. 

We are especially interested in rapid updates, because, by limits imposed by this timescale on how much information can be gathered by journalists, these updates are more likely to contain single units of information, updates and quotes. Thus, in our experiments, we focus on \textit{The New York Times}, \textit{Independent}, \textit{Associated Press}, \textit{Washington Post}, and \textit{BBC}. We also include \textit{Guardian} and \textit{Reuters} because they typically compete directly with the previously mentioned outlets in terms of content and style, even if they do not publish as frequently. 

\subsubsection{Discourse Across Time}

We are interested in the dynamics of articles over time. Although this analysis is still ongoing, we seek to understand how, as the article grows through time, the types of information included in it changes.We show in Figure \ref{fig:discourse_dynamics:dist_by_len} and \ref{fig:discourse_dynamics:odds_by_version} that in later versions and longer articles\footnote{Version Number has spearman's correlation $r=.335$ with article length.} sentences are dominated by \textit{Distant} discourse.

Interestingly, later versions are also more likely to have \textit{Main} and \textit{Cause} discourse added. Based on our annotator interviews, we surmise that this is because, for breaking news, a journalist is frequently trying to assess the causes behind the story. In early drafts, we also see \textit{Main} sentences being removed. This is due to, as the story is updating in early versions, the \textit{Main} event is most likely to be changing.

\subsubsection{Top Words}

\begin{table}[t]
    \centering
    \small
    \begin{tabular}{l|p{5cm}}
    \toprule
    Unchanged &  said, trump, people, president, concerns, government,  year \\
    \hline
    Add/Del &  says, senate, law, death, wednesday, monday, tuesday \\
    \bottomrule
    \end{tabular}
    \caption{Top Words in Additions/Deletions vs. top words in unchanged sentences.}
    \label{tab:top_words}
\end{table}

\noindent\textbf{Top Words:} We characterize added and deleted sentences by their word usage in Table \ref{tab:top_words}. Words indicating present-tense, recent updates are more likely: day-names like ``Monday'' or ``Tuesday'' and the present-tense verb ``says'' (compared with the past-tense ``said'' in unchanged sentences).

\subsubsection{Collection of Corrections, Authorship}
\label{app:corrections}

To identify instances of \textit{Corrections} in added sentences, we used the following lexicon:

``was corrected'', ``revised'', ``clarification'', ``earlier error'', ``version'', ``article''

Here are some examples of corrections:

\begin{itemize}
    \item CORRECTION: An earlier version of this story ascribed to Nato spokesman Brig Gen Carsten Jacobsen comments suggesting that after Saturday\'s shooting, people would have to be ``looking over their shoulders'' in Afghan ministries.
    \item CORRECTION 19 November 2012:An earlier version of this story incorrectly referred to ``gargoyles'', not ``spires''.
    \item Correction 7 March 2012: An earlier version of this story mistakenly said Rushbrook's car had been travelling at 140mph at the time of the crash.
\end{itemize}

To identify instances of \textit{Contributor Lines}, we use the following lexicon:

``reporting by'', ``additional reporting'', ``contributed reporting'', ``editing by''

Here are some examples of contributor lines:

\begin{itemize}
    \item Additional reporting by Simon Browning.
    \item 'The article relied heavily on reporting by Reuters and the BBC, and it cited Reuters in saying that during a visit in October 1989 by Pope John Paul II to South Korea, China had prevented the pope’s airplane from flying through Chinese airspace.
    \item The revelation comes after reporting by The New York Times last week showing that the head of communications at the N.I.H.’s parent agency, the Department of Health and Human Services, also accused federal scientists of using the coronavirus to try to defeat Mr. Trump.
    \item Additional reporting by Daniel Strauss in Richmond, Virginia, Richard Luscombe in West Palm Beach, Florida, and Ed Pilkington in Essex Junction, Vermont.
\end{itemize}


\begin{table*}[t]
\centering
\small
\begin{tabular}{|l|r|r|l|l|l|l|l|}
\hline
Source & \# Articles & \# Versions &  Start & End & Ctry. & Lang.  & Coll. \\
\hline
    BBC &       307,616 & 1,244,490 & 2006-08  &  2021-01 &    U.K. &  En. & NS \\
    Guardian &  231,252 &   852,324 &  2012-01 &  2021-01 &    U.K. &  En. & NS \\
    Nytimes &   87,556 &    395,643 &  2012-08 &  2020-12 &    U.S. &  En. & NS \\
    Telegraph & 78,619 &    124,128 &  2017-01 &  2018-09 &    U.K. &  En. & NS \\
    Fox &       78,566 &    117,171 &  2017-01 &  2018-09 &    U.S. &  En. & DE \\
    CNN &       58,569 &    117,202 &  2017-01 &  2018-09 &    U.S. &  En. & DE \\
    Independent & 55,009 &  158,881 &  2014-01 &  2018-05 &    U.K. &  En. & NS \\
    CBC &         54,012 &  387,292 &  2017-08 &  2018-09 &  Ca. &  En. & DE \\
    Dailymail &   50,639 &  166,260 &  2017-01 &  2018-09 &    U.K. &  En. & DE \\
    BBC &         42,797 &  99,082 &  2017-01 &  2018-09 &     U.K. &  En. & DE \\
    La Presse &    40,978 &  73,447 &  2017-08 &  2018-09 &  Ca. & Fr-Ca. & DE \\
    Torontostar & 33,523 &  310,112 &  2017-08 &  2018-07 &    Ca. & En. &  DE \\
    Globemail &   32,552 &  91,820 &  2017-08 &  2018-09 &     Ca. & En. &  DE \\
    Reuters &     31,359 &  143,303 &  2017-01 &  2018-09 &    U.K. &  En. & DE \\
    National Post & 22,934 & 63,085 &  2017-08 &  2018-09 &    Ca. &  En. & DE \\
    Associated Press & 22,381 & 97,314 &  2017-01 &  2018-09 &  U.S. & En. & DE \\
    Washington Post &  19,184 & 68,612 &  2014-01 &  2020-07 &   U.S. & En. &  NS \\
    Toronto Sun &      19,121 & 46,353 &  2017-08 &  2018-09 &   Ca. &  En. & DE \\
    Calgary Herald &   7,728 &  33,427 &  2017-08 &  2018-09 &   Ca. & En. &  DE \\
    The Rebel &        4,344 &  19,383 &  2017-08 &  2018-09 &  Ca. &  En. &  DE \\
    Canada Land &      65 &     101 &  2017-12 &  2018-09 &    Ca. &  En. & DE \\
\hline
\end{tabular}
\caption{A summary of the number of total number of articles and versions for different media outlets which comprise our dataset. Also shown is the original collection that they were derived from (DE for DiffEngine, and NS from NewsSniffer), and the date-ranges during which articles from each outlet were collected.}
\label{tbl:source_list}
\end{table*}

\subsection{Dataset Tables and Fields}

Our dataset is released in a set of 5 SQLite tables. Three of them are primary data tables, and two are summary-statistic tables. Our primary data tables are: \texttt{articles}, \texttt{sentence\_diffs}, \texttt{word\_diffs}; the first two of which are shown in Tables \ref{tbl:layout_articles} and \ref{tbl:layout_sdiffs} (\texttt{word\_diffs} shares a similar structure with \texttt{sentence\_diffs}). We compile two summary statistics tables to cache statistics from \texttt{sentence\_diffs} and \texttt{word\_diffs}; they calculate metrics such as \texttt{NUM\_SENTENCES\_ADDED} and \texttt{NUM\_SENTENCES\_REMOVED} per article.\footnote{These summary statistic tables make it convenient to, say, filter \texttt{sentence\_diffs} in order train a model on all articles that have one sentence added; or all articles that have no sentences removed.}

The \texttt{sentence\_diffs} data table's schema is shown in Table \ref{tbl:dbschemas} and some column-abbreviated sample rows are shown in Table \ref{tbl:demos}. As can be seen, the diffs are calculated and organized on a sentence-level. Each row shows a comparison of sentences between \textit{two adjacent versions of the same article.}\footnote{So, for instance, article A, with versions 1, 2 where each version has sentences i, ii, iii, would have 3 rows (assuming sentences were similar): A.1-2.i, A.1-2.ii, A.1-2.iii.} Every row in \texttt{sentence\_diffs} contains index columns: \texttt{SOURCE}, \texttt{A\_ID}, \texttt{VERSION\_OLD}, and \texttt{VERSION\_NEW}. These columns can be used to uniquely map each row in \texttt{sentence\_diffs} to \textit{two} rows in \texttt{article}.\footnote{One mapping for \texttt{sentence\_diffs.VERSION\_OLD} = \texttt{article.VERSION\_ID} and one mapping for \texttt{sentence\_diffs.VERSION\_NEW} = \texttt{article.VERSION\_ID}.} 

\subsection{\texttt{TAG} columns in \texttt{sentence\_diffs}}

The columns \texttt{TAG\_OLD} and \texttt{TAG\_NEW} in \texttt{sentence\_diffs} have specific meaning: how to transform from version to its adjacent version. In other words, \texttt{TAG\_OLD} conveys where to find \texttt{SENT\_OLD} in \texttt{VERSION\_NEW} and whether to change it, whereas \texttt{TAG\_NEW} does the same for \texttt{SENT\_NEW} in \texttt{VERSION\_OLD}. 

More concretely, consider the examples in Table \ref{tbl:demo1}, \ref{tbl:demo2} and \ref{tbl:demo3}. As can be seen, each tag is 3-part and has the following components. \textbf{Component 1} can be either \textbf{M}, \textbf{A}, or \textbf{R}. \textbf{M} means that the sentence in the current version was \textbf{M}atched with a sentence in the adjacent version, \textbf{A} means that a sentence was \textbf{A}dded to the new version and \textbf{R} means the sentence was \textbf{R}emoved from the old version.\footnote{i.e. an \textbf{A}dded row is not present in the old version and a \textbf{R}emoved row is not present in the new version. They have essentially the same meaning and we could have condensed notation, but we felt this was more intuitive.} \textbf{Component 2} is only present for \textbf{M}atched sentences, and refers to the index or indices of the sentence(s) in the adjacent version\footnote{I.e. in \texttt{TAG\_OLD}, the index refers to the \texttt{SENTENCE\_ID} of \texttt{SENT\_NEW}}. Additionally, \textbf{Component 3} is also only present if the sentence is \textbf{M}atched. It can be either \textbf{C} or  \textbf{U}. \textbf{C} refers to whether the matched sentence was \textbf{C}hanged and \textbf{U} to whether it was \textbf{U}nchanged.

Although not shown or described in detail, all \textbf{M} sentences have corresponding entry-matches in \texttt{word\_diffs} table, which has a similar schema and tagging aim.

A user might use these tags in the following ways:

\begin{enumerate}
    \item \label{itm:firstusecase} To compare only atomic edits, as in \newcite{faruqui2018wikiatomicedits}, a user could filter \texttt{sentence\_diffs} to sentences where \textbf{M..C} is in \texttt{TAG\_OLD} (or equivalently, \texttt{TAG\_NEW}). Then, they would join \texttt{TAG\_OLD.Component\_2} with \texttt{SENTENCE\_ID}. Finally, they would select \texttt{SENT\_OLD}, \texttt{SENT\_NEW}.\footnote{or simply look in the \texttt{word\_diffs} table.}
    \item To view only refactorings, or when a sentence is moved from one location in the article to another, a user could filter \texttt{sentence\_diffs} to only sentences containing \textbf{M..U} and follow a similar join process as in use-case \ref{itm:firstusecase}.
    \item\label{itm:thirdusecase} To model which sentences might be added, i.e. $p(\text{sentence}_i \in \text{article}_{t+1} | \text{sentence}_i \nin \text{article}_t)$, a user would select all sentences in \texttt{SENT\_OLD}, and all sentences in \texttt{SENT\_NEW} where \textbf{A} is in \texttt{TAG\_NEW}.
    \item To model the inverse of use-case \ref{itm:thirdusecase}, i.e. which sentences would be removed, or $p(\text{sentence}_i \nin \text{article}_{t+1} | \text{sentence}_i \in \text{article}_t)$, a user would select all sentences in \texttt{SENT\_NEW}, and all sentences in \texttt{SENT\_OLD} where \textbf{R} is in \texttt{TAG\_OLD}.
\end{enumerate}

\begin{table*}[t]
\centering
\subfloat[DB schema for the \texttt{article} table. \texttt{SOURCE}, \texttt{A\_ID} and \texttt{VERSION\_ID} are the primary key columns. \label{tbl:layout_articles}]{
\begin{tabular}{|p{2.9cm}|p{1cm}||p{2.9cm}|p{1cm}||p{2.9cm}|p{1cm}|}
\hline
         Column Name &     Type  & Column Name &     Type  &  Column Name &     Type  \\
\hline
    SOURCE & index &      TITLE & text & CREATED & text \\
    A\_ID &  index &      URL &  text &  ARCHIVE\_URL &   text \\
    VERSION\_ID & index & TEXT & text &  NUM\_VERSIONS &  int \\
\hline
\end{tabular}
} \\
\subfloat[DB schema for the \texttt{sentence\_diffs} table (\texttt{word\_diffs} is similar). Table compares \textit{version pairs} of articles. The rows in the table are on the sentence-level; \texttt{V\_OLD\_ID} refers to the index of the old version, \texttt{V\_NEW\_ID} refers to the index of the new version. \texttt{TAG\_OLD} gives information for how to transition from the old version to the new version; \texttt{TAG\_NEW} is the inverse.\label{tbl:layout_sdiffs}]{
\begin{tabular}{|p{2.9cm}|p{1cm}||p{2.9cm}|p{1cm}||p{2.9cm}|p{1cm}|}
\hline
   Column Name &     Type  & Column Name &     Type  &  Column Name &     Type  \\
\hline
    SOURCE &     index & V\_NEW\_ID &    index &  TAG\_OLD &     text \\
    A\_ID &      index & SENTENCE\_ID &  index &  SENT\_NEW &     text \\
    V\_OLD\_ID & index & SENT\_OLD &     text &   TAG\_NEW &     text \\
\hline
\end{tabular}}
\caption{Schemas for two databases central to our content organization scheme.}
\label{tbl:dbschemas}
\end{table*}

\subsection{Comparison With Other Edits Corpora}
\label{app:related_work}
Here, we give a tabular comparison with other edits corpora, showing our 

\begin{table*}[t]
    \centering
    \begin{tabular}{|p{2cm}|p{4cm}|p{1.5cm}|p{2cm}|p{3cm}|}
    \hline
        Corpus & \# Revisions & Language & Source & Goal \\
    \hline
    \hline
        WiKed Error Corpus
        &  12 million changed sentences 
        & English
        & Wikipedia
        & Grammatical Error Correction (GEC) \\
        \hline
        WikiAtomic-Edits
        & 43 million ``atomic edits''\footnote{An ``atomic edit''}
        & 8 languages 
        & Wikipedia
        & Language Modeling \\
        \hline
        WiCoPaCo
        & 70,000 changed sentences
        & French
        & Wikipedia
        & GEC and Sentence paraphrasing \\
        \hline
        WikiHow-ToImprove
        & 2.7 million changed sentences
        & English
        & WikiHow
        & Version prediction, article improvement \\
        \hline
        NewsEdits
        & 36.1 million changed sentences, 21.7 million added sentences, 14.2 million removed sentences. 72 million atomic edits.
        & English and French
        & 22 media outlets
        & Language modeling, event sequencing, computational journalism \\
    \hline
    \end{tabular}
    \caption{A comparison of natural langauge revision history corpora.}
    \label{tab:corpora_comparison}
\end{table*}

\section{Algorithm Details}
\label{app:alg_details}

In this section, we give further examples further justify our asymmetrical sentence-matching algorithm. The examples shown in Tables \ref{tbl:demo1}, \ref{tbl:demo2} and \ref{tbl:demo3} illustrate our requirements. The first example, shown in Table \ref{tbl:demo1}, occurs when a sentence is edited syntactically, but its meaning does not change\footnote{Syntactic changes: synonyms are used, or phrasing is condensed, but substantially new information is not added}. So, we need our sentence-matching algorithm to use a sentence-similarity measure that considers semantic changes and does not consider surface-level changes. The second example, shown in Table \ref{tbl:demo2}, occurs when a sentence is split (or inversely, two sentences are merged.) Thus, we need our sentence matching algorithm to consider many-to-one matchings for sentences. The third example, shown in Table \ref{tbl:demo3}, occurs when sentence-order is rearranged, arbitrarily, throughout a piece. Finally, we need our sentence-matching algorithm to perform all pairwise comparisons of sentences. 

\begin{table*}[t]
\small
\centering
\subfloat[
    Demo 1: Word-Level atomic edit corrections applied when a sentence-level match is found, using the \texttt{difflib} Python library.
    \label{tbl:demo2}
]{
\begin{tabular}{|p{.4cm}|p{.4cm}|p{5.6cm}|p{5.6cm}|p{.4cm}|} 
\hline
Sent Idx & Old Tag & Old Version & New Version & New Tag\\
\hline
1 & M 1 C & 
\cellcolor{pink}The Bundesbank would only refer to an interview \colorbox{DarkPink}{Mr.} \colorbox{DarkPink}{Weidmann} \colorbox{DarkPink}{gave} \colorbox{DarkPink}{to} Der Spiegel magazine last week, in which \colorbox{DarkPink}{he} said, ``I can \colorbox{DarkPink}{do} my \colorbox{DarkPink}{job} best \colorbox{DarkPink}{by} \colorbox{DarkPink}{staying} in office.'' &  
\cellcolor{LimeGreen}The Bundesbank would only refer to an interview \colorbox{DarkLimeGreen}{published} \colorbox{DarkLimeGreen}{in} Der Spiegel magazine last week, in which \colorbox{DarkLimeGreen}{Mr.} \colorbox{DarkLimeGreen}{Weidmann} said, ``I can \colorbox{DarkLimeGreen}{carry} \colorbox{DarkLimeGreen}{out} my \colorbox{DarkLimeGreen}{duty} best \colorbox{DarkLimeGreen}{if} \colorbox{DarkLimeGreen}{I} \colorbox{DarkLimeGreen}{remain} in office.'' & 
M 1 C \\
\hline
\end{tabular}
}
\\
\vspace{.1cm}
\subfloat[
    Demo 2: A sentence that is split results in the addition of a new sentence, but is matched with the previous dependent clause. Minimal word-level edits are applied.
    \label{tbl:demo1}
]{
\small
\begin{tabular}{|p{.4cm}|p{.4cm}|p{5.6cm}|p{5.6cm}|p{.4cm}|} 
\hline
Sent Idx & Old Tag & Old Version & New Version & New Tag\\
\hline
1 & M 1 2 C & 
\cellcolor{pink}DALLAS—Ebola patient Thomas Eric Duncan told his fiancee the day he was diagnosed last week that he regrets exposing her to the deadly virus \colorbox{DarkPink}{and} \colorbox{DarkPink}{had} he known he was carrying Ebola, he would have “preferred to stay in Liberia and died than bring this to you,” a family friend said & 
\cellcolor{LimeGreen} DALLAS—Ebola patient Thomas Eric Duncan told his fiancee the day he was diagnosed last week that he regrets exposing her to the deadly virus\colorbox{DarkLimeGreen}{.}  & 
M 1 \\
\hline
2 & {} & {} & \cellcolor{LimeGreen} \colorbox{DarkLimeGreen}{Had} he known he was carrying Ebola, he would have “preferred to stay in Liberia and died than bring this to you,” a family friend said. & 
M 1 C \\
\hline
\end{tabular}
}
\\
\vspace{.1cm}
\subfloat[Demo 3: Two features shown: (1) Refactoring, or order-swapping, makes sentences appear as though they have been deleted and then added. Swapped sentences are matched through their tags. (2) The last sentence is a newly added sentence and is not matched with any other sentence.\label{tbl:demo3}]{
\begin{tabular}{|p{.4cm}|p{.4cm}|p{5.6cm}|p{5.6cm}|p{.4cm}|} 
\hline
Sent Idx & Old Tag & Old Version & New Version & New Tag\\
\hline
1 & M 2 U & \cellcolor{pink} ``The mother, this was the first time seeing her son since he got to the States. & 
\cellcolor{LimeGreen} ``She has not seen him for 12 years, and the first time she saw him was through a monitor,'' said Lloyd. & M 2 U \\
\hline
2 & M 1 U & \cellcolor{pink} She has not seen him for 12 years, and the first time she saw him was through a monitor,'' said Lloyd. &
\cellcolor{LimeGreen} ``The mother, this was the first time seeing her son since he got to the States.''
& M 1 U \\
\hline
3 & {} & {} &
\cellcolor{LimeGreen} ``She wept, and wept, and wept.''
& A \\
\hline
\end{tabular}
}
\vspace{.1cm}
\caption{Here we show demos of three tricky edge-cases and how our tagging scheme handles them. 
\texttt{Old Tag} annotates a \texttt{Old Version} relative to changes in the \texttt{New Version} (or ``converts'' the \texttt{Old Version} to the \texttt{New Version}). \texttt{New Tag} is the inverse. Tag components: \textbf{Component 1: M, A, R.} Whether the sentence is \textbf{M}atched, \textbf{A}dded, or \textbf{R}emoved. \textbf{Component 2: Index.} If \textbf{M}atched, what is the index of the sentence in version that it is matched to. \textbf{Component 3: C, U.} If \textbf{M}atched, is the sentence \textbf{C}hanged or \textbf{U}nchanged.}
\label{tbl:demos}
\end{table*}

\begin{algorithm}[t]
\SetAlgoLined
\SetKwInOut{Input}{input}
\SetKwInOut{Output}{output}
\Input{Article versions $v_{old}$, $v_{new}$, Match Threshold $T$}
\Output{maps $m_{old \rightarrow new}$, $m_{old \leftarrow new}$}
initialize;\\
$m_{old \rightarrow new}$, $m_{old \leftarrow new}$ = \{\}, \{\};\\
 \tcp{match $v_{old} \rightarrow v_{new}$}
 \For{$(i, s_i) \in v_{old}$}{
    $d = \max_{s_j \in v_{new}} \text{Sim}_{asym}(s_i, s_j)$\\
    $j = \argmaxA_{s_j \in v_{new}} \text{Sim}_{asym}(s_i, s_j)$\\
    $m_{old \rightarrow new}\left[i\right] = j \times \mathbbm{1} \left[d > T\right]$\\
 }
 \tcp{match $v_{old} \leftarrow v_{new}$}
 \For{$(j, s_j) \in v_{new}$}{
    $d = \max_{s_i \in v_{old}} \text{Sim}_{asym}(s_j, s_i)$\\
    $i = \argmaxA_{s_i \in v_{old}} \text{Sim}_{asym}(s_j, s_i)$\\
        $m_{old \leftarrow new}\left[j\right] = i \times \mathbbm{1} \left[d > T\right]$
    
 }
\caption{Asymmetrical sentence-matching algorithm. Input $v_{old}$, $v_{new}$ are lists of sentences, and output is an index mapper. If a sentence maps to 0 (i.e. $d < T$), there is no match. $Sim_{asym}$ is described in text.}
\label{alg:matching}
\end{algorithm}

 \subsection{Refactors}
 \label{app:alg_details:refactor}
 
To identify which sentences were \textit{intentionally} moved rather than moved as a consequence of other document-level changes, we develop an iterative algorithm based on the idea that a refactor is an intentional sentence movement that creates an edge-crossing. Algorithm \ref{alg:refactor} givens our algorithm.

In English, our algorithm represents sentence matches between two article versions as a bipartite graph. We use a Binary Tree to recursively find all edge crossings in that graph. This idea is based off of the solution for an SPOJ challenge problem: \url{https://www.spoj.com/problems/MSE06H/}\footnote{Solution given here: \url{https://github.com/akhiluanandh/SPOJ/blob/master/MSE06H.cpp}.}. We extend this problem to return the \textit{set} of all edge crossings, not just the crossing number.

Then, we filter edge crossings to a candidate set, applying the following conditions in order and stopping when there is only one edge crossing left: (1) edges that have the most number of crossings (2) edges that extend the most distance or (3) edges that move upwards. In most cases, we only apply the first and then the second conditions. In very rare cases, we apply all three. In rarer cases, we apply all three and \textit{still} have multiple candidate edges. In those cases, we just choose the first edge in the candidate set. We continue removing edges until we have no more crossings.

\begin{algorithm*}[t]
\SetAlgoLined
\SetKwInOut{Input}{input}
\SetKwInOut{Output}{output}
\Input{Sentence matches, i.e. edges $e$ between doc $i$ and doc $j$, as a list of tuples: $e_i = (s_{i1}, s_{i2}), e_j = (s_{j1}, s_{j2})...$.}
\Output{Minimal set of edges $r$ that, when removed, eliminate all crossings.}
\tcp{Subroutine identifies all edge crossings in $e'$ and returns mapping $c = \{ e_i \rightarrow [e_j, e_k...] , e_j \rightarrow ...\}$ from each edge to all its crossings.}
$c = getEdgeCrossings(e)$ \\
\While {$|c| > 0$}{
    \tcp{Find candidate set: all edges with maximum crossings.}
    $m = \max_i |c[e'_i]|$\\
    $e' = e'_i $ where $|c[e'_i]| = m$\\
    \uIf {$|e'| > 1$} {
        \tcp{Filter candidate set: all edges $\in e'$ that extend the maximum distance.}
        $d = \max_i |e'_i[0] - e'_i[1]| $ \\
        $e' = e'_i $ where $|e'_i[0] - e'_i[1]| = d$\\
        \uIf{$|e'| > 1$}{
            \tcp{Filter candidate set: all edges $\in e'$ that move up.}
            $e' = e'_i $ where $e'_i[1] - e'_i[0] < 0$\\
            }
        }
    }
    \tcp{Take first element of $e'$ as the candidate to remove.}
    $t = e'[0]$ \\
    $r.push(t)$ \\ 
    \tcp{Remove $t$ from $c$ and from all $c[e'_i]$ lists that contain it.}
    $c = removeEdge(t)$
\caption{Identifying Refactors. We define refactors as the minimal set of edge crossings in a bipartite graph which, when removed, remove all edge crossings.}
\label{alg:refactor}
\end{algorithm*}

\section{Annotation-Task Descriptions}
\label{app:annotation}

\subsection{Task: Sentence Matching}

We give our annotators the following instructions:

\begin{quote}
The goal of this exercise is to help us identify sentences in an article-rewrite that contain substantially new information. To do this, you will identiy which sentences match between two versions of an article.

Two sentences match if:
\begin{enumerate}
    \item They are nearly the same, word-for-word.
    \item They convey the same information but are stylistically different.
    \item They have slightly different information but have substantial overlap in meaning and narrative function.
\end{enumerate}
Examples of Option 3 include (please see the ``Examples'' section for real examples):
\begin{enumerate}
    \item Updating events.
    \begin{itemize}
        \item (Ex) The man was presumed missing. → The man was found in his home.
        \item (Ex) The death count was at 23. → 50 were found dead.
        \item (Ex) The senators are still negotiating the details. → The senators have reached a deal.
    \end{itemize}
    \item An improved analysis.
    \begin{itemize}
        \item (Ex) The president is likely seeking improved relations. → The president is likely hoping that hard-liners will give way to moderates, improving relations.
        \item (Ex) The storm, a Category IV, is expected to hit Texas. → The storm, downgraded to Category III, is projected to stay mainly in the Gulf.
        \item (Ex) Analysts widely think the shock will be temporary. → The shock, caused by widespread shipping delays, might last into December, but will ultimately subside.
    \end{itemize}
    \item A quote that is very similar or serves the same purpose.
    \begin{itemize}
        \item (Ex) ``We knew we had to get it done.'' said Senator Murphy. → ``At the end of the day, no one could leave until we had a deal'' said Senator Harris.
        \item (Ex) ``It was gripping.'' said the bystander. → ``I couldn't stop watching.'' said a moviegoer.
    \end{itemize}
\end{enumerate}
Two sentences do not match if:
\begin{enumerate}
    \item They contain substantially different information.
    \item They serve different narrative functions.
    \item There is a much better match for one sentence somewhere else in the document.
\end{enumerate}
Things to keep in mind:
\begin{itemize}
    \item Two sentences might match even if they are in different parts of the document.
    \item One sentence can match with multiple other sentences, because that sentence might be split up into multiple sentences, each with similar information as parts of the original.
    \item Sentences don't have to match.
    \begin{itemize}
        \item Substantially new information, perspectives or narrative tools might be added in a new version.
        \item Substantially old information, perspectives or narrative tools might be removed from an old version.
    \end{itemize}
\end{itemize}
\end{quote}

\begin{figure*}[t]
    \centering
    \includegraphics[width=\linewidth]{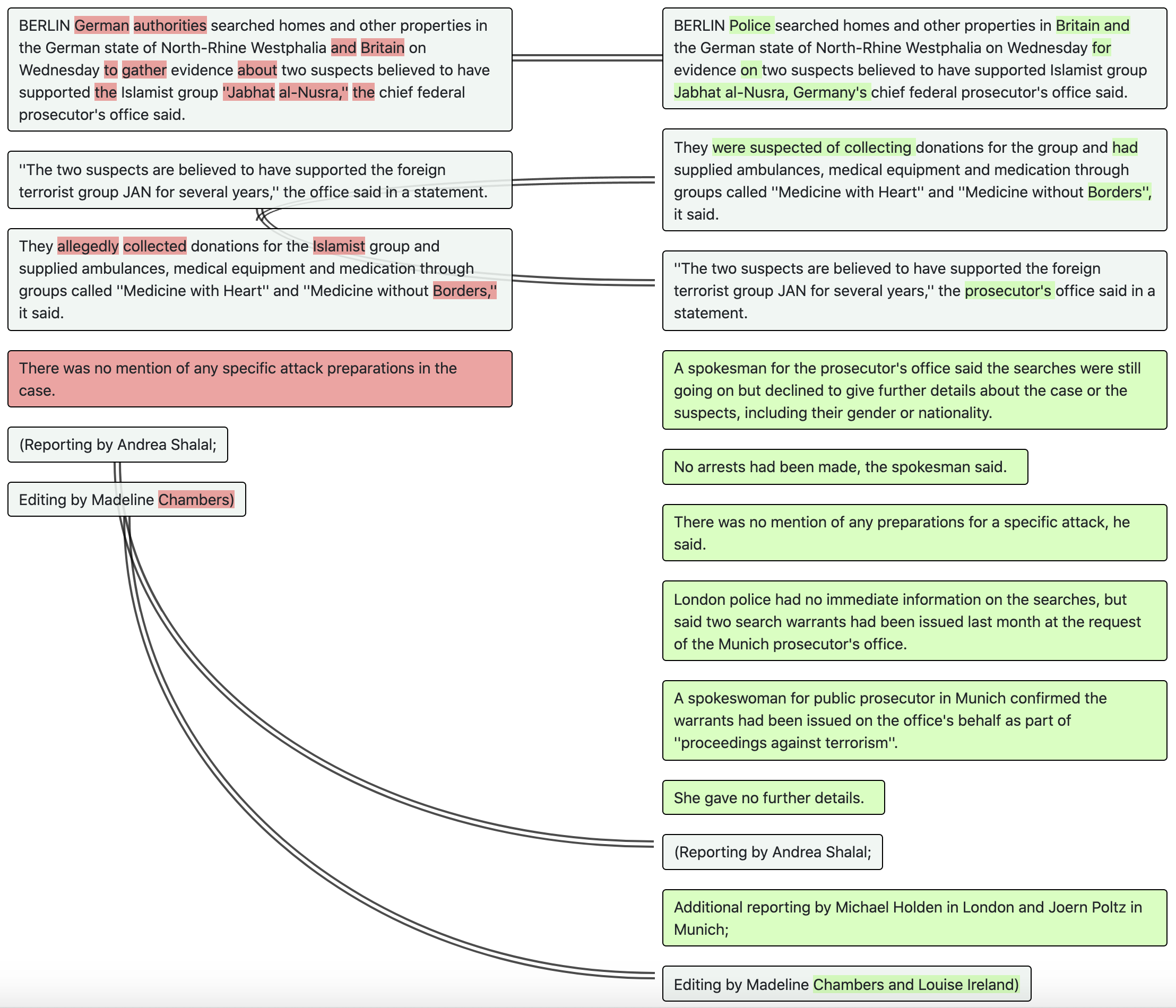}
    \caption{Example of Sentence Matching Task. All lines represent sentences that have been matched. When the user hits ``Submit'', additional coloring is added to the unmatched sentences, which represent \textit{Addition} (green, right) and \textit{Deletion} (red, left) sentences.}
    \label{fig:mturk_task_1}
\end{figure*}

Annotators completed the task by drawing lines between sentences in different versions of an article. An example is shown in Figure \ref{fig:mturk_task_1}. We use highlighting to show when non overlapping sequences in the inbox, using simple lexical overlap. If the user mouses over a text block, they can see which words do no match between all textblocks on the other side. Although this might bias them towards our lexical matching algorithms, we do not see them beaking \textbf{TB-medium}. This was very helpful for reducing the cognitive overload of the task.

\begin{figure*}[t]
    \centering
    \includegraphics[width=\linewidth]{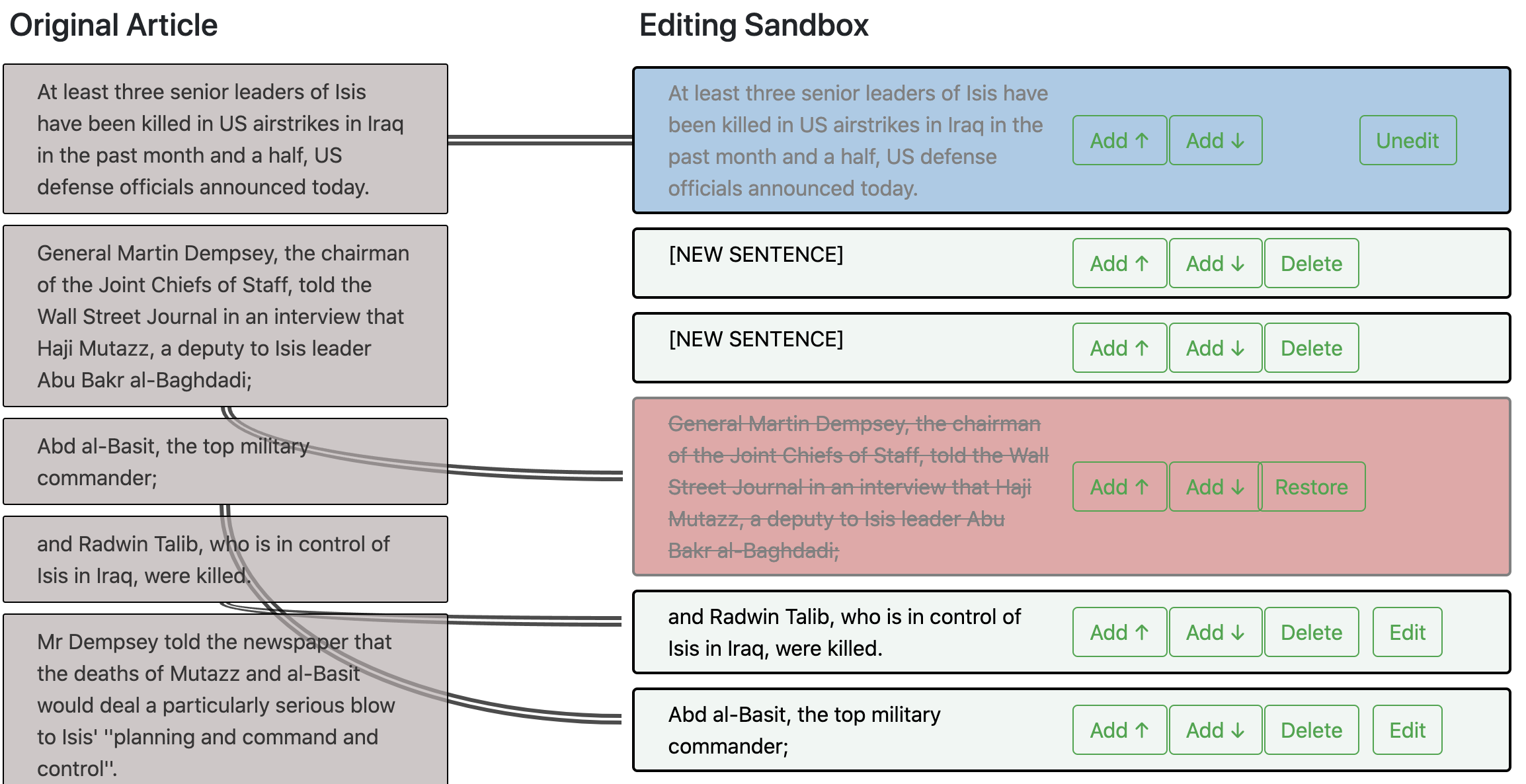}
    \caption{Example of Editing Task. The gray boxes on the left serve as a reference for how the original article was written. The sandbox on the right is where annotators actually perform the task. The first sentence has been \textit{Edited}, two sentences have been \textit{Added}, the third has been \textit{Deleted} and the fourth has been \textit{Refactored} downwards.}
    \label{fig:mturk_task_2}
\end{figure*}

\subsection{Task: Edit Actions}

In this task, workers were instructed to perform edit operations to an article version in anticipation of what the next version would look like. We recruited 5 workers: journalists who collectively had over a decade of experience working for outlets like \textit{The New York Times}, \textit{Huffington Post}, \textit{Vice}, a local outlet in Maine, and freelancing.

We gave our workers the following instructions.

\begin{quote}
You will be adding, deleting and moving sentences around in a news article to anticipate what a future version looks like.

\begin{itemize}
    \item \textbf{Add a sentence either below or above the current sentence} by pressing the Add ↑ or Add ↓ buttons. Adding a sentence means that you feel there is substantially new information, a novel viewpoint or quote, or necessary background information that needs to be present.
    \item \textbf{Move a sentence by dragging it around on the canvas.} Moving a sentence, (or what we're calling refactoring) means that the importance of a sentence should be either increased or decreased within the article. Please note: refactors are rare!
    \item \textbf{Delete a sentence by hitting the Delete button.} Deleting an Added sentence just reverses that action---we will not record this. Deleting a sentence that is present means you feel it needs to be (a) substantially rewritten (ergo: a new sentence should also be Added), or (b) the sentence no longer applies given new information that was added.
    \item \textbf{Edit a sentence by hitting the Edit button.} Editing a sentence means that the wording might change a little bit due to other changes happening around the sentence or events within the sentence being updated.    
    \item \textbf{Leaving a sentence unchanged} means that you don't really expect the sentence to change at all in the next version of the article.
\end{itemize}
When you're ready to submit, please hit the Submit button and please check to see what the actual edits were so you can improve for next task!
\end{quote}

\subsection{Annotator Analysis}

\begin{table}[t]
    \small
    \centering
    \begin{tabular}{llr}
    \toprule
    Worker Id &  Num Tasks Completed \\
    \midrule
    ASQL7ZBXI7WF6 &                  101 \\
    A2E8P5A3IKROKB &                   92 \\
    A17GX84A96WF6C &                   31 \\
    A1685VEOIJIUMR &                   13 \\
    A2USH7VYFMU1ME &                    5 \\
    A30BGCC8EC1NW &                    3 \\
    \bottomrule
    \end{tabular}
    \caption{Count of Tasks Completed per worker}
    \label{tab:worker_task_count}
\end{table}

\begin{table}[t]
    \centering
    \small
    \begin{tabular}{lr}
    \toprule
         Worker Id &  Accuracy Across Tasks \\
    \midrule
    A2E8P5A3IKROKB & 76.6 \\
    A30BGCC8EC1NW & 58.3 \\
    ASQL7ZBXI7WF6 & 46.0 \\
    A17GX84A96WF6C & 38.7 \\
    A2USH7VYFMU1ME & 35.0 \\
    A1685VEOIJIUMR & 30.8 \\
    \bottomrule
    \end{tabular}
    \caption{Accuracy across document tasks (i.e. \% bins correct across document-level subtasks: \textit{Added}, \textit{Edited}, \textit{Deleted}, \textit{Refactored}).}
    \label{tab:worker_accuracy}
\end{table}

\begin{figure}[t]
    \centering
    \includegraphics[width=\linewidth]{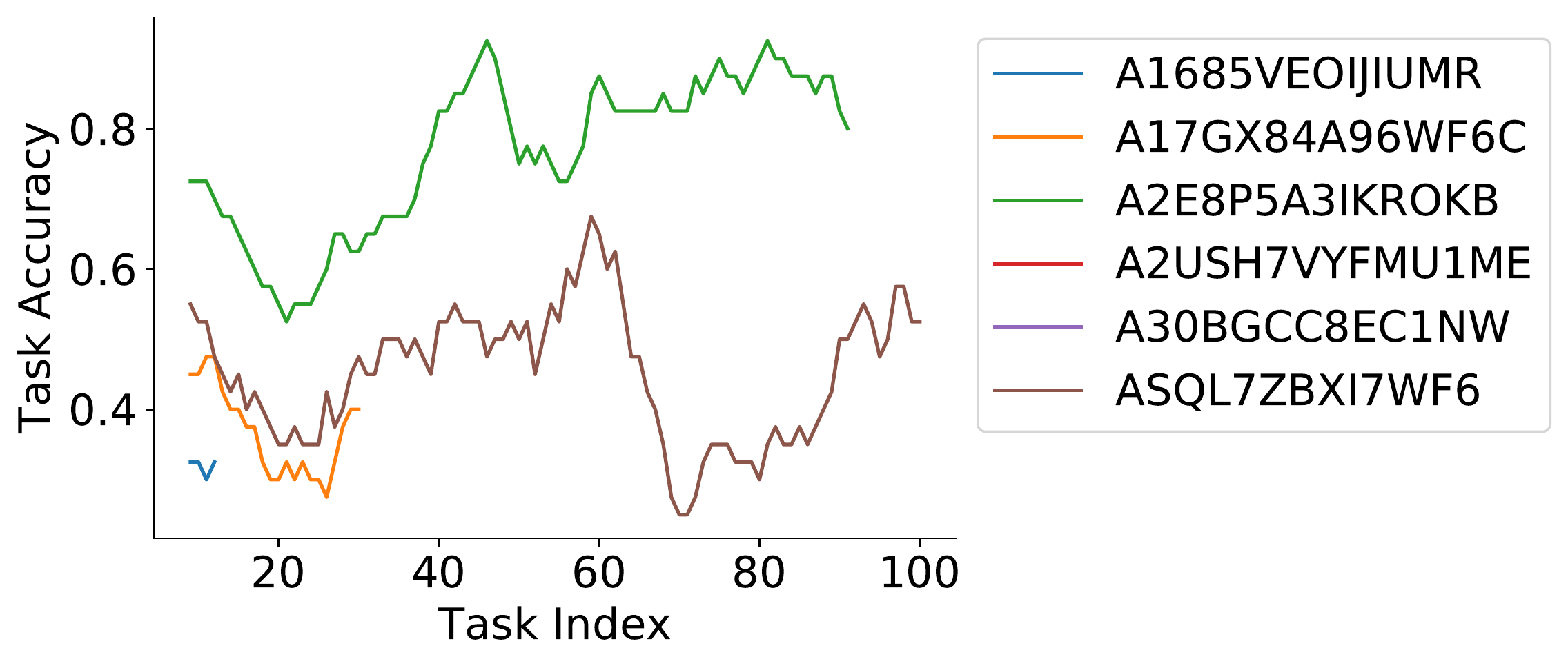}
    \caption{Worker Accuracy over time, by task}
    \label{fig:worker_acc_over_time}
\end{figure}

We seek here to characterize the performance of different expert annotators. We see in Table \ref{tab:worker_task_count} that there are three workers which do over $30$ tasks each. We characterize the per-task accuracy by counting the number of edit-operations per document, and seeing if they got the same number as the true number of edits (each expressed as a binned count i.e. low: $[0,1)$ operations, medium: $[1,3)$ operations, high: $[3, \infty)$ operations).

We show that there is a wide variety of performances, in Table \ref{tab:worker_accuracy}, with some workers getting over 75\% of the operations correct and others getting $\approx 30\%$ correct.

Interestingly, we see that there is a learning process occurring. In Figure \ref{fig:worker_acc_over_time}, we see that workers get better over time as they do more tasks. This indicates that the training procedure of letting them see the edits that actually happened is successful at teaching them the style and patterns the edits will take.

\subsection{Annotator Interview 1}
\label{app:annotation:interview_1}
This annotator was involved in the Editing task. They edited 50 stories.

\begin{enumerate}
\item \textit{What was your general thought process?}
    \setlength{\parindent}{\enumerateparindent}
    Well, my first general though was: ``how do I do this update?'' Then I thought back to the instructions, and really tried to predict how the AP\footnote{The AP, or \textit{The Associated Press}, sets many standards for journalistic writing and reporting cycles.} would update.
	
	I then had to decide what timespan I'd use---in general, I assumed a 24 hour update window, but sometimes it was different. If the story updates 2 hours after news breaks vs. 2 days, it will look very different
	
	Sometimes, I would read the story, try to figure out what the story was about, ask what was missing, what I'd include in a story if I was reporting it fully. A lot of times what I felt were missing were more causal analysis, more quotes, more perspectives.
	
	As I was going through, I almost always decided to edit the lede, and was almost always correct with that. Most leads, I thought, could be more efficient, they could incorporate more details from further down in the story into the lede. Also, as stories unfolded, the actor responsible for the event becomes clear, that information will get added to the lede. For example, a building collapses in Manhattan -> faulty beam causes the building collapse. This detail often only becomes apparent afterwards.
	
	What I realized doing this was that there are different genres of breaking news article, and genre matters a lot for how it gets updated. These are the following categories:
	\begin{enumerate}
    	\setlength{\parindent}{\enumerateparindent}
	
	    \item Stories where the future is contingent, and you're making predictions in realtime.

        ex) A sailor went missing off the isle of Mann. This story is fundamentally about an unknown -- will he be discovered or not? This is one of the harder ones to figure out how to update. How it plays out determines how it will be updated. If the search goes on for a long time, you'll have more details, you'll have quotes from his family, conditions on the water. If he's found, this stuff becomes irrelevant. You'll have information about how he gets found, then you'll have information about how many people get updated.
	    
	    ex) A story was about ``Trump is about to make a speech''. ``Trump expected to speak''. I updated it as if event didn't happen yet. But the real update actually contained him speaking. Stories about when multiple futures can happen, without knowing the timescale of the update, are difficult to predict.
	        
	    I determined whether an event was unfolding by looking for several clues. I looked for certain words: ``expected'', ``scheduled'', etc. Usually this signals an event-update. I looked for stories where there's a ton of uncertainty. 
        
        Another clue was that the only sources are official statements (ex. ``Officials in Yemen say something happened''.) The space of possible change increases. You're going to get conflicting reports, eye-witnesses contradicting official statements.
        
        Some articles included direct appeals to readers---``don't use the A4 if you're traveling between London, etc.'' For crime articles: ``if you have any information, please contact agency.'' This kind of direct appeal is not relevant in the next version.
	    
	    \item Past stories when the event is totally in the past.
        
        For these stories, I looked for vagueness of the original article to determine what would be updated. If it's more specific, for example, with exact death toll numbers, information about specific actors and victims, the less it's going to be updated. For these stories, my tendency was to add at least 1-2 sentences of context towards the end of every story. If you're writing for Reuters, you might not need that. 
        
	    In general, I wanted to see some background, people involved.
	    
	    The quotes you're getting, are they press releases or are they directly from people? If they more official statements and press releases, then you'll see more updates in the form of specific victim quotes. 
    \end{enumerate}
	
	One general note: most breaking stories were about bad things. Disasters, crashes, missing people, etc. For a bombing, there's a pretty predictable pattern of expansion. Death toll will get added, more eyewitness accounts. It has an expansionary trajectory.
	
\item \textit{How did you determine if a sentence needed to be added?}
	I decided to add anywhere I saw vagueness. I added a lot towards the beginning, right after the nut graf is where I added the most sentences. If I saw a sentence taken from a press release, I added after that, assuming that the journalist would get a more fleshed-out quote from someone. 
	
	Often I added [sentences] at the end to add context. I never added something before the lead. 
    
    Maybe a story has two ideas, then I'd add sentences to the second half to flesh out a second idea. 
    
    Sometimes I thought about different categories of information---quotes, analysis, etc.---and it was obvious if some of that was missing.
    
\item \textit{How did you determine if a sentence needed to be deleted? }

    I very rarely thought things needed to be deleted
	
	One of the challenges of the experiment was that it was hard to indicate how to combine sentences. I got around this by hitting ``edit'' for sentences that needed to be combined. Then I'd delete ones below, assuming that the edited sentence would include a clause from the sentence below it. 
	
	Structural sentences and cues got deleted often. Sentences like ``More follows'', etc. Nothing integral to the substance of the story.
	
	I noticed that almost always, [informational content of sentences that had been deleted] had been reincorporated.

\item \textit{How did you determine if a sentence needed to be moved up/down?}

    I did this by feel, what seemed important. One example: A building collapse in Morocco. A sentence way towards the end had a report about weak foundations, that needed to be brought up. This indicated that the journalist became more confident about something
	
	The inverted pyramid so widely used, in a breaking news it's fairly easy to weight the importance of different elements. Thus, I rarely felt the need to move items upwards.
	
	Sometimes I saw examples of when what was initially a small quote from official was expanded in a later version. Then, it was brought up because the quote became more important. But usually, my instinct would not be to move quotes from officials up.
	
\item \textit{Did it help to see what actually happened after you finished the task?}

    Usually there was 1-2 things that we had done that were basically the same. 
    
    A couple of times, [I] was satisfied to see that the updated story made the same decision to switch sentences around.

\item \textit{Any general closing thoughts?}
    
    Most interesting thing was to see how formally constrained journalists and editors are, and how much these forms and genres shape your thought and your work.
    
    There are assumptions get baked into the genres about who's credible, what kinds of things carry weight, sorts of outcomes deserve special attention, a whole epistemic framework. 
    
    Even though there's a lot of variation, there's a fair amount of consistency. 
    
    I was disappointed that, especially for rapidly expanding stories, the edits were mainly causes and main events. I saw very few structural, causal analyses added to breaking stories. There was some analysis that got added to one story about bombings in the Middle East, but still, not a whole lot about how the specific conflict originated.
\end{enumerate}

\subsection{Annotator Interview 2}
\label{app:annotation:interview_2}
This annotator was involved in both the editing task and the version-prediction task. They annotated over 100 examples of the first task, and 50 of the second.

\begin{enumerate}
    \setlength{\parindent}{\enumerateparindent}
    \item \textit{What was your general thought process while doing the edits task?}
	
	First, before starting, I made the assumption that every story would need edits, because I think everything could always use more work. In reality, if the article wasn't updated the way it was, I was representing one option. My process was:
	\begin{enumerate}
	    \item Read the whole story, don't make any changes at first.
	    \item Then, I would think about what I thought was the most important sentence.
	    \item I would often pull that high up into the lede, and then I'd add a sentence before or after.
	\end{enumerate}
	
	The factors that determined the most important part of the article were:
    \begin{enumerate}
        \setlength{\parindent}{\enumerateparindent}
        \item Some indication of harm done or the most recent development. I always took ``harm done'' as the most important part of a story.
        
        For example: Death count---20 people were killed in some explosion vs. a bomb went off here. Moved the ``20 people killed'' higher because that was a harm
    		ex. Officials are investigating whether so-and-so doctored documents. 
    		
    	\item Then, I would add/delete and edit based on these. So, I would create a new sentence and edit the next sentence to give more context.
    \end{enumerate}
    
    \item \textit{How did you determine if a sentence needed to be added?}
    
	So, after identifying the lede that I described previously, I went through and looked through what parts I felt needed more context or a quote. Getting quotes was very important. Often I identified events that I thought warranted a reaction, acknowledgment, information from a source. If these weren't there, I added a sentence. I didn't keep a checklist of these elements (i.e. ``quote'', ``context'', etc.) It was more a gut feeling about what it needed. If I were going back and doing it again, I would write out a checklist.

	Often, especially when the news was unpredictable, I would often add a sentence in the beginning saying ``I don't know what this sentence is going to be, but it's going to be something''. In other words, I was adding context to what the unknown would be. I was able to do this pretty successfully, to predict what context would happen around the unpredictable event.

	Where I tried to add more information to flesh out certain unknowns:
	\begin{enumerate}
	    \item If an official said something that needed to be followed up on, I would delete these and add new sentences
	    \item I had hoped that the reporter would get that information themselves through eyewitnesses, court documents, etc. 
	    \item Sometimes an official would give filler quotes like: ``we'll have more information later this afternoon''. These would be replaced with the actual update.
	    \item Context: I would add historical context. How often has something been occurring in this area, etc. Many of real updates did have these contextual sentences. 
	\end{enumerate}

    \item \textit{How did you decide whether a sentence needed to be edited?}
    
	After I decided what would be moved up, I looked at details (dates, people, etc.). Sentences with details were the ones that were most likely to be edited.

\item \textit{ How did you determine if a sentence needed to be deleted? }

	I deleted sentences that were redundant. I identified filler quotes (e.g. officials saying they'll get more information soon.). These would be deleted when, presumably, more information did come in. Sometimes a quote was redundant to a sentence that was already there. One of the challenges was deciding when to delete or edit a sentence.

\item \textit{How did you determine if a sentence needed to be moved up/down?}

	I almost always moved sentences upwards, to the top. As we discussed previously, the top then needs to have room for an update. Again, as we discussed previously, I used harm and recent developments as a metric to decide where to move. The context was also moved around based on when the events took place.

	I also tried to focus on recent developments. For example: ``Officials are investigating whether so-and-so doctored documents''. I would move that to the top. I pulled up the active part of the article to express what was actually happening.

\item \textit{What things did you get wrong?}
	
	I was really bad at predicting stories that were ``delete all'', ``replace all''. I struggled more with stories that were about political leaders speaking at an event or speaking at a conference, because these ended up going different ways. Sometimes they made a big announcement that would make headlines, but it was hard to known beforehand what that announcement would be.

	For crime, or spot news, it was clearer that an event was unfolding and would have specific updates. By ``spot news'', I mean stories about crimes, fires, rescues, weather events/disasters, etc. -- something unexpected as opposed to articles about events that have been planned, like the example of a political figure speaking at a conference. It was these unexpected events that actually follow more predictable paths when they unfold.
	
	I saw a lot of discrepancies between sentences I chose to edit, and then the actual result was that they got deleted. For example, the death toll was in a sentence, and I'd edit that sentence, but they chose to add a sentence with the same information. The sentence matching algorithm didn't do a good job with informational units that were not at the sentence level.

\item \textit{ How did you assess uncertainty in an article?}

	Often it was topic-based. I can't think of key indicators that I used to assess uncertainty.

\item \textit{Was really helpful after I made the edits to see what actually happened?}

	I tried to balanced this with what my natural instincts were. I did get better over time. I did feel more confident over time. The changes would be more in my decisions to edit vs. add/delete. In my head, I had the same end result in mind, but they edited it and I added a new sentence. I never felt I was widely off 	

\item\textit{ Did you see a lot of analytical pieces? Or mainly breaking news?}

	I saw a mix of stuff that was analytical vs. factual. There were certainly more breaking news events, events that were going to happen and change on the same day. However, I did see some day 2 stories. Sometimes, they were updates that were part of an ongoing investigation. The breaking stories and spot news, crime, were the easiest to do. Those ones seem much more formulaic.

\item \textit{ What was your general thought process while doing the versioning task? How did you identify versions that updated? }

This one was trickier because I would assume that everything would be updated, everything would be improved. The mindset change that I made was ``Will this story itself be edited, or will they write a followup with more information?'' Once I made this separation this became easier 
	
\item \textit{What patterns did you observe in this task?}

The timing of when I thought an update would occur ended up mattering a lot. I paid closer attention to stories that would have updates within the same day or a short period of time. The longer the time-periods between updates, the more likely a new piece would be published instead of an update.
	
Again, crime and spot news it was clear — the person was on the scene at this minute, they'd get more information. 
	
The other giveaways were ``so and so is expected to deliver remarks later this afternoon.'' It wasn't quite a preview of the event but it would clearly be updated
	
The other thing that made me choose to mark a story as ``would be updated'' is if there was a key perspective missing or if there was no quotes at all. By ``key perspective'', I mean, a key quote from a participant that is usually present in this type of story. For a crime, for example, this included: Law enforcement perspective, witness, family. In general, it means that both sides are represented.

\item \textit{Were there examples that you thought would update that didn't?}

	There were some with stock figures, quarterly earnings, that I initially thought would be updated, but I had seen the examples that were filled out, but I'd be more accepting that this was a final report and that it's not going to have any quotes. I became better at identifying which types of pieces wouldn't have context or quotes. 	

\item\textit{ Anything I may have missed?}

I tried to flag a couple of articles that transferred over inaccurately. Sometimes there were cases of where one article published to the same URL was something completely different. Sometimes there were calls for subscribing to newsletters or related story links. I deleted ones that were repetitive. This might have influenced results on some articles. These structural updates were annoying.

\item \textit{Could you see solving this kind of prediction task as being useful in a newsroom? }

	I could see it being used as a people management tool. Newsrooms are desperate for any kind of methodology to guide the decisions they make. Deciding who should attack a new story, and who should stay put working on their old piece would help a lot!

\end{enumerate}

\end{document}